# Feature Based Methods in Domain Adaptation for Object Detection: A Review Paper

Helia Mohamadi [0009-0009-9942-0024], MohammadAli Keyvanrad [0000-0002-7654-1001] and MohammadReza Mohammadi [0000-0002-1016-9243]

[1] Faculty of Electrical & Computer Engineering Malek Ashtar University of Technology Tehran, Tehran, Iran
[2] Faculty of Electrical & Computer Engineering Malek Ashtar University of Technology Tehran, Tehran, Iran
[2] School of Computer Engineering, Iran University of Science and Technology (IUST), Tehran, Iran

Helia.mohamadi@outlook.com, keyvanrad@mut.ac.ir, mrmohammadi@iust.ac.ir

**Abstract.** Domain adaptation, a pivotal branch of transfer learning, aims to enhance the performance of machine learning models when deployed in target domains with distinct data distributions. This is particularly critical for object detection tasks, where domain shifts—caused by factors such as lighting conditions, viewing angles, and environmental variations—can lead to significant performance degradation. This review delves into advanced methodologies for domain adaptation, including adversarial learning, discrepancy-based, multi-domain, teacher-student, ensemble, and Vision Language Models techniques, emphasizing their efficacy in reducing domain gaps and enhancing model robustness. Feature-based methods have emerged as powerful tools for addressing these challenges by harmonizing feature representations across domains. These techniques, such as Feature Alignment, Feature Augmentation/Reconstruction, and Feature Transformation, are employed alongside or as integral parts of other domain adaptation strategies to minimize domain gaps and improve model performance. Special attention is given to strategies that minimize the reliance on extensive labeled data and using unlabeled data, particularly in scenarios involving synthetic-to-real domain shifts. Applications in fields such as autonomous driving and medical imaging are explored, showcasing the potential of these methods to ensure reliable object detection in diverse and complex settings. By providing a thorough analysis of state-of-the-art techniques, challenges, and future directions, this work offers a valuable reference for researchers striving to develop resilient and adaptable object detection frameworks, advancing the seamless deployment of artificial intelligence in dynamic environments.

## 1 Introduction.

Object detection plays a fundamental role in various machine learning applications, from autonomous driving to medical imaging. However, a persistent challenge remains: models trained in one domain often underperform when applied to another due to domain shifts. Domain shifts or domain gaps are the variations in data distributions between source and target domains (Tanveer et al. 2023). This issue is particularly pronounced when models trained on synthetic data are deployed in real-world scenarios, leading to notable performance degradation. For instance, a model trained on high-resolution images from a DSLR camera may perform poorly on images captured by a smartphone due to differences in image quality and characteristics.

Domain adaptation, a subset of transfer learning, seeks to mitigate this problem by aligning the discrepancies between source and target domains. Feature-based domain adaptation methods focus on harmonizing feature representations across domains to improve detection accuracy. Techniques such as adversarial feature alignment (Zheng et al. 2020; Weng and Yuan 2024; Kay et al. 2024) have demonstrated effectiveness in achieving this goal. Additionally, approaches like unified multi-granularity alignment (Zhang et al. 2024b) and sequence feature alignment (Wang et al. 2022a) have been proposed to enhance the robustness of object detection systems.

This review examines state-of-the-art feature-based domain adaptation techniques, categorizing them into discrepancy-based, adversarial-based, and multi-domain-based approaches. It explores their applications and effectiveness in addressing domain shifts in object detection, highlighting key research trends such as the integration of deep learning frameworks and the use of synthetic-real data blending. Furthermore, the review identifies ongoing challenges, including computational costs and the risk of negative transfer, and discusses potential solutions to advance the field.

By providing a comprehensive overview of current methodologies and their applications, this paper aims to contribute to the development of robust and adaptable object detection systems capable of performing effectively across diverse environments.

In transfer learning and domain adaptation, dealing with both a source and a target domain can lead to shifts in data distribution, potentially impacting model performance. As illustrated in **Fig. 1** Domain shifts can be classified into four distinct categories (Kouw and Loog 2018) as covariate shift, label shift, conditional shift, and concept shift, each influencing adaptation strategies to optimize model efficacy. This paper thoroughly examines each shift, discussing its mathematical foundations, examples, and the resulting implications for model architecture and adaptation techniques.

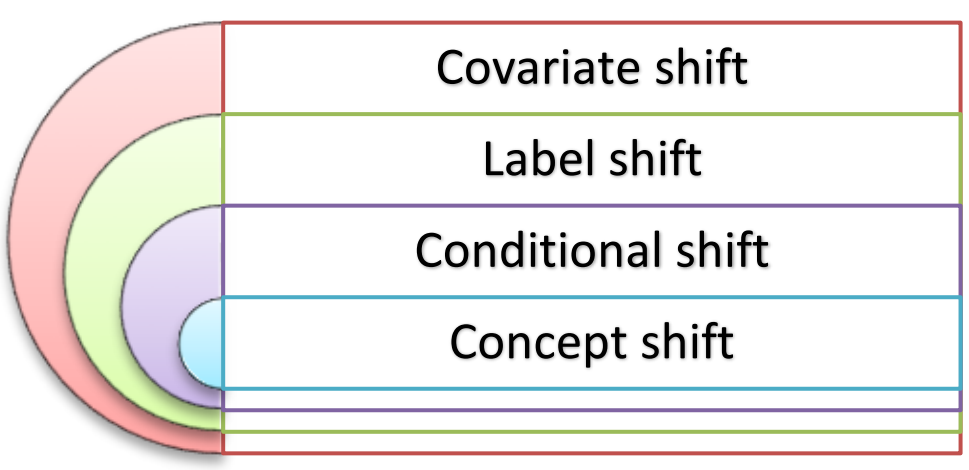

**Fig. 1.** summary of four possible domain shifts

**Covariate Shift: Covariate Shift:** Covariate shift occurs when the distribution of input features changes between the source and target domains, while the conditional distribution of the output given the input remains constant. This shift can be addressed using feature probability matching techniques, which align the feature distributions across domains (Wen et al. 2023). This means that $P_{source}\,(X) \neq P_{target}(X)$, but $P_{source}\,(Y|X) = P_{target}(Y|X)$. For example, a model trained on images of cats in bright lighting may perform poorly on images of cats in dim lighting due to changes in feature distribution (Xu et al. 2024a). Covariate shift can decrease model accuracy due to mismatched feature distributions, despite stable input-output relationships. Addressing this shift typically involves reweighting or normalizing features to achieve alignment across domains.

**Label Shift:** Label shift, also referred to as target shift or prior shift, occurs when the label distribution varies between the source and target domains, though the conditional feature distribution given labels remain the same. This situation necessitates methods like class probability matching, which focuses on adjusting the model to account for variations in class probabilities rather than feature distributions (Park et al. 2023; Wen et al. 2023). This means that $P_{source}\,(Y) \neq P_{target}(Y)$, but $P_{source}\,(X|Y) = P_{target}(X|Y)$. For example, a model trained to classify emails as spam or not may face challenges if the proportion of spam emails increases in the target domain. To address that, Class probability matching (CPM) is proposed to effectively estimate class probability ratios under label shifts (Wen et al. 2023).

**Conditional Shift:** Conditional shift refers to variations in the conditional distribution of labels given the input features, even if marginal distributions of features and labels remain unchanged. This means that $P_{source}\,(X) = P_{target}(X)$, also $P_{source}\,(Y) = P_{target}(Y)$, but $P_{source}\,(Y|X) \neq P_{target}(Y|X)$. In sentiment analysis, certain phrases or words can convey different sentiments across domains; for instance, "fine" might indicate a positive sentiment in one context and a neutral one in another. Conditional shift can be challenging to manage, as it changes the feature-label association directly. Solutions may involve domain-specific adjustments or retraining to adapt to target-specific requirements (Park et al. 2023; Wen et al. 2023).

**Concept Shift (Concept Drift):** Concept drift, also termed concept shift, represents shifts in data distributions over time that can degrade the predictive performance of machine learning models. This phenomenon typically arises from variations in the joint distribution $P(X,Y)$, which may manifest as covariate drift—changes in the feature distribution P(X)—or as label drift, involving shifts in the conditional distribution $P(Y|X)$. Such shifts may result from external factors like evolving user behavior or inherent fluctuations in the data, making the detection and adaptation to concept drift vital for model resilience in dynamic settings (Galmeanu and Andonie 2024; Lukats et al. 2024). Several methods exist to detect concept drift, ranging from supervised approaches to unsupervised strategies that are particularly beneficial in cases of limited labeled data. Techniques like KL divergence are frequently used to assess distributional changes, while real-time methods such as DriftLens leverage deep learning to identify drift both quickly and accurately (Greco et al. 2024). For adaptive responses, models may employ continuous kernel learning or ensemble approaches to dynamically adjust weights and maintain performance as data evolves (Chen and Dai 2024). Privacy-preserving models further enhance these adaptations by safeguarding sensitive information in high-stakes applications (Varshney and Torra 2024).

In this review, we explore the recent advancements in feature-based domain adaptation methods, which play a crucial role in improving the generalization of object detection models across various environments. These methods aim to reduce the distribution gap between the source and target domains, thus enabling more effective knowledge transfer. As object detection is increasingly applied in real-world scenarios, such as autonomous driving, medical imaging, and surveillance, addressing domain shifts has become an essential challenge.

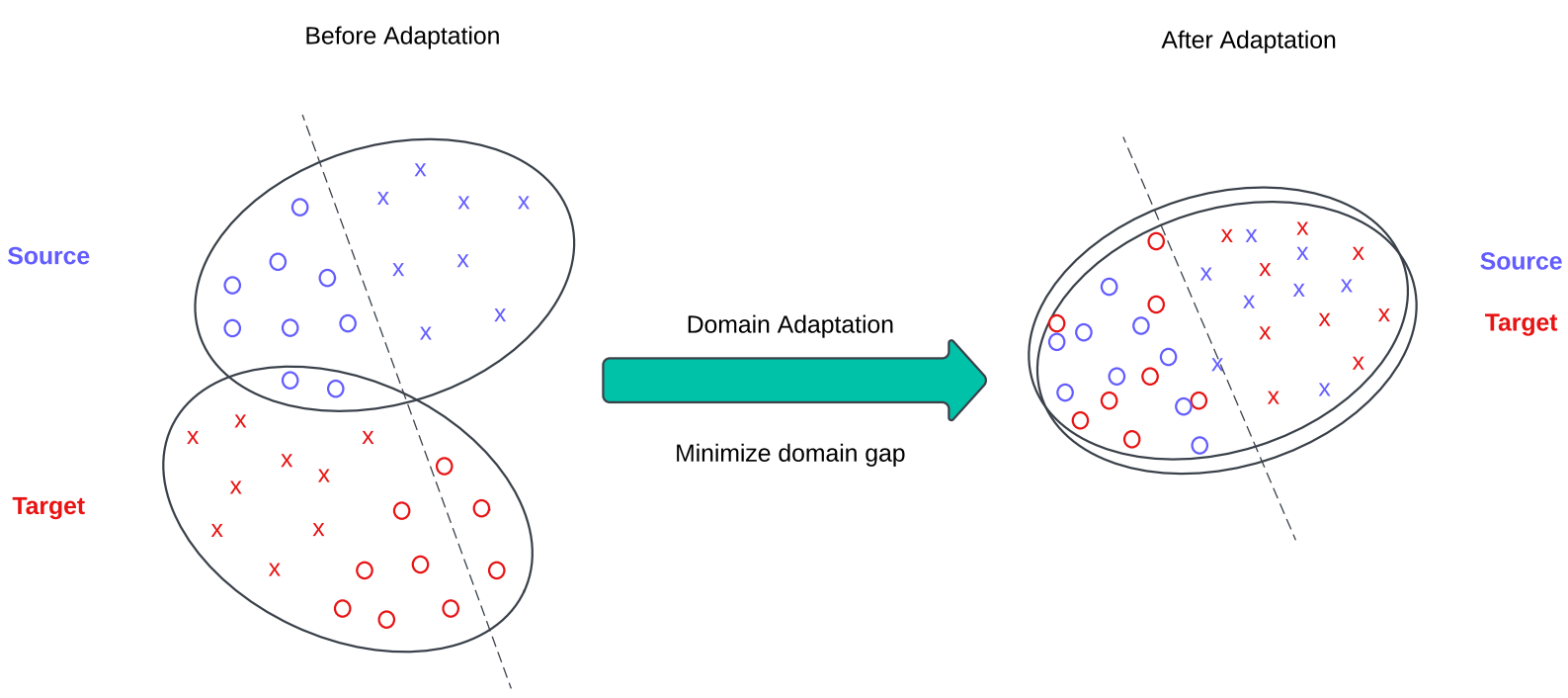

**Fig. 2.** Illustration of domain adaptation reducing the domain gap between source and target domains. Before adaptation, the source (blue) and target (red) data distributions exhibit a significant domain gap, causing misalignment in feature space. After adaptation, domain adaptation methods harmonize feature distributions, facilitating improved model performance across domains.

In this review paper, we provide a comprehensive analysis of the key methodologies and frameworks utilized in domain adaptation for object detection. The remainder of this article is organized as follows: In Section 2.1, we categorize domain adaptation methods based on the level of supervision, including supervised, unsupervised, semi-supervised, and weakly-supervised approaches. Section 2.2 delves into object detection architectures, examining one-stage, two-stage, and hybrid models. Section 2.3 focuses on the core domain adaptation strategies, covering discrepancy-based, adversarial-based, teacher-student-based, ensemble-based, and vision-language model (VLM)-based methods. In Section 2.4, we turn our attention to feature-based adaptation approaches, exploring techniques such as feature alignment, feature augmentation/reconstruction, and feature transformation. Section 3 outlines the key datasets commonly employed to benchmark and evaluate domain adaptation methods for object detection. Validation metrics are discussed in Section 4, providing insights into how the effectiveness of these methods is quantified. Finally, Section 5 synthesizes the findings by summarizing the progress made in the field and identifying potential avenues for future research. This structured review aims to serve as a valuable reference for researchers by highlighting the critical components and advancements in domain adaptation for object detection. The overall structure of this review is outlined in **Fig. 3**.

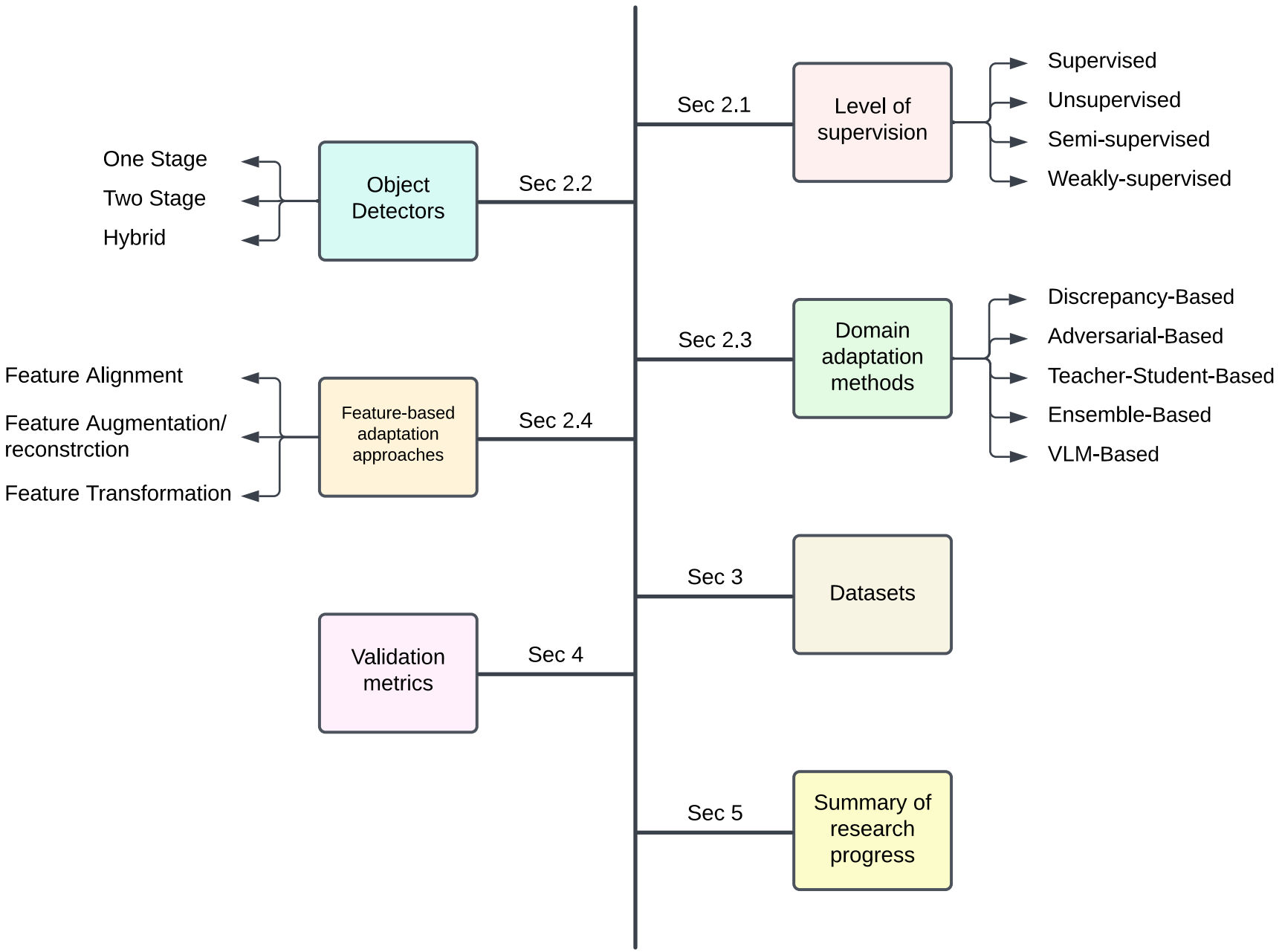

**Fig. 3.** Conceptual framework of this review paper.

## 2 Methodologies and Frameworks

In this section, we delve into the specific methods and techniques employed in feature-based domain adaptation for object detection, as outlined in the previous overview. These methods aim to address the challenges posed by domain shifts, which can significantly degrade the performance of machine learning models when transferred from a source to a target domain. We categorize the various approaches based on their underlying strategies, including discrepancy methods, adversarial methods, multi-domain methods, teacher-student methods, ensemble methods, and vision-language model methods. Each of these categories represents a distinct approach to mitigating domain adaptation challenges, such as feature distribution mismatches and label inconsistencies. In the following sections, we examine the theoretical foundations of each method, explain their implementation, and discuss their applications in the context of object detection. This methodology provides a comprehensive framework for understanding the progression and current state of feature-based domain adaptation techniques, highlighting both their strengths and limitations in real-world applications.

### 2.1 Level of Supervision

Domain adaptation is an essential technique within the realms of machine learning and computer vision, aimed at facilitating the transfer of knowledge from a well-defined source domain to a target domain that may be deficient in labeled data. This process is particularly crucial in enhancing model performance in novel and uncharted environments, thereby alleviating the burdens associated with extensive and often expensive data labeling efforts. The effectiveness of domain adaptation is significantly influenced by the type and level of supervision available during the adaptation process, which can be classified into four primary categories: supervised, semi-supervised, weakly supervised, and unsupervised approaches. Each category presents unique challenges and strategies that impact the overall success of the adaptation process in real-world applications.

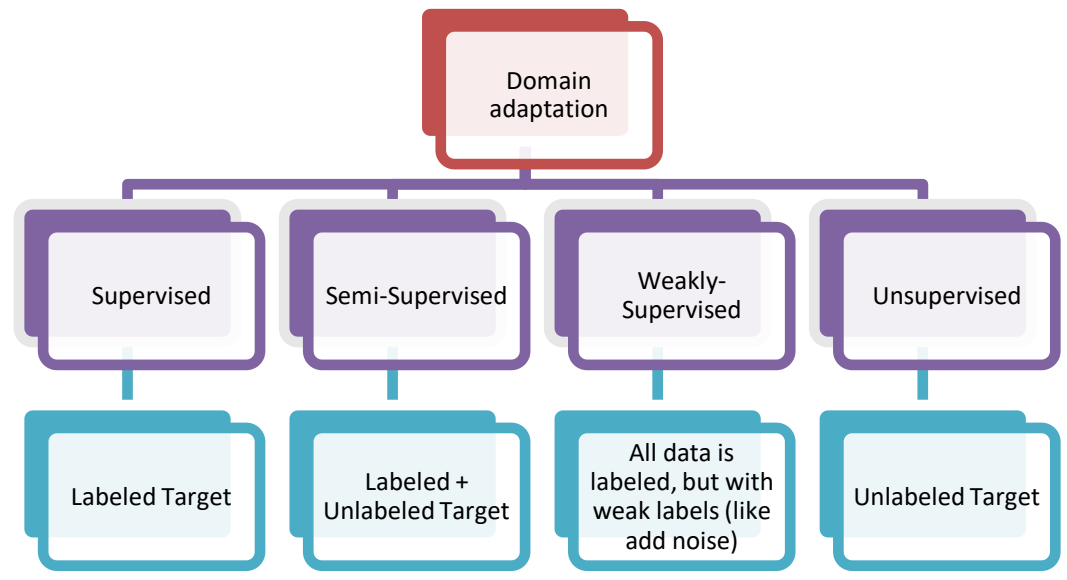

**Fig. 4.** Categorization of Domain Adaptation Methods by Level of Supervision

**Supervised domain adaptation (SDA)** addresses the challenge of transferring a model that has been trained on a source domain to a target domain, wherein variations in data attributes—such as illumination, feature distributions, or methodologies of data acquisition—may result in suboptimal performance of the model. The primary objective of SDA is to alleviate these discrepancies through the fine-tuning of the model with labeled data sourced from the initial domain, while concurrently utilizing unlabeled data from the target domain, thereby enhancing generalization across disparate domains. This methodology has been demonstrated to be efficacious in fields such as computer vision, natural language processing, and medical imaging, wherein the capacity to adapt to novel data environments is of paramount importance. Nevertheless, substantial challenges persist, particularly in managing significant domain shifts and optimizing adaptation under conditions of limited labeled data from the target domain. Consequently, ongoing scholarly inquiry is devoted to the exploration of advanced methodologies, to bolster the performance of SDA including adversarial learning and self-supervised techniques. Furthermore, SDA exhibits conceptual connections with related domains such as domain generalization and out-of-distribution detection, indicative of a broader initiative to enhance model adaptability within diverse and dynamic real-world contexts (Xu et al. 2019; Liu et al. 2022; HassanPour Zonoozi and Seydi 2023).

**Semi-Supervised Domain Adaptation (SSDA)** extends the existing domain adaptation framework by tackling scenarios wherein the target domain is characterized by a deficiency of labeled data. In SSDA, a model that has been trained on a source domain with a wealth of labeled data is adapted for performance within a target domain that, while exhibiting similarities, possesses distinctive characteristics and a scarcity of labeled data. The principal challenge in SSDA lies in the effective utilization of this limited dataset from the target domain. This challenge is mitigated through methodologies that align feature representations across domains, such as those predicated on minimizing conditional entropy—wherein the model retains uncertainty regarding unlabeled data while incentivizing the feature encoder to acquire discriminative features. The Minimax Entropy (MME) approach has surfaced as a preeminent solution within this domain, demonstrating considerable enhancements in SSDA tasks. This technique offers a robust mechanism for adapting models with minimal labeled data, yielding significant practical ramifications across various sectors of machine learning (Xu et al. 2019; Saito et al. 2019; Yu and Lin 2023; Hosseini and Caragea 2023).

**Weakly Supervised Domain Adaptation (WSDA)** further broadens the domain adaptation paradigm to contexts wherein the data from the target domain is both limited and characterized by noise. The WSDA framework is specifically designed to adapt a model from a well-annotated source domain to a target domain where annotations are sparse and potentially unreliable. To accomplish this, WSDA utilizes techniques such as Transferable Curriculum Learning (TCL) (Shu et al. 2019a), which identifies reliable examples from the source domain for transfer, thereby mitigating the influence of noisy data from the target. TCL facilitates the prioritization of meaningful information from the source domain, thereby enhancing the model's capacity to generalize to the target domain. This methodology is particularly advantageous in practical applications where pristine, well-labeled data is not readily accessible, enabling models to derive insightful conclusions from noisy or incomplete datasets. TCL has exhibited superior efficacy in WSDA tasks, presenting a pragmatic solution for the adaptation of models to challenging environments beset with imperfect data (Shu et al. 2019b; Lan et al. 2024).

**Unsupervised Domain Adaptation (UDA)** addresses the intricate challenge of adapting a machine learning model that has been trained on labeled data from a source domain for application in a target domain that is devoid of labeled data. To elucidate, the source domain may be compared to a meticulously organized library of annotated texts, while the target domain can be likened to a repository of cryptic, untranslated manuscripts. The principal aim is to exploit the knowledge acquired from the annotated texts to comprehend the manuscripts, notwithstanding the lack of direct translations. UDA methodologies concentrate on harmonizing feature representations across different domains, mitigating domain variances, and employing strategies such as adversarial learning and self-training to enhance model adaptation and facilitate the learning process from unlabeled data. This approach enables the model to extract substantial insights from the target domain, even in the absence of labeled instances, analogous to a literary sleuth deciphering the enigmas of an obscure language. Recent advancements in UDA have demonstrated its effectiveness, outpacing conventional techniques and underscoring its capacity to unveil latent knowledge across a variety of applications (Miller 2019; Xu et al. 2022; Liu et al. 2022).

### 2.2 Object Detection

Domain adaptation plays a critical role in object detection, especially when models trained on source need to be effectively applied to a target without significant loss in performance. This is essential because in real-world applications, the source and target data often differ significantly in terms of lighting conditions, camera angles, backgrounds, or even object appearances. For example, an object detection model trained on clear, high-resolution daytime images may perform poorly on low-light, nighttime images or when deployed in different environments like urban versus rural settings. The ability to adapt a model to new domains is vital to ensure the robustness and generalizability of object detection systems across diverse scenarios. In autonomous driving, for instance, the vehicle must accurately detect objects like pedestrians and vehicles regardless of weather conditions or time of day. Similarly, in medical imaging, models trained on one type of equipment or image quality should perform equally well when applied to images from different sources or hospitals.

Domain adaptation techniques help address these challenges by either minimizing the domain shift between the source and target data or by enabling the model to learn domain-invariant features. This can significantly improve the detection accuracy in the target domain without needing expensive, large-scale labeled data from every possible domain variation. In this way, domain adaptation enhances the flexibility and applicability of object detection models in real-world, dynamic environments, making it a crucial aspect of advancing AI and computer vision technologies.

The accomplishment of object detection relies on a range of technologies, including deep neural networks like YOLO[1] (Redmon et al. 2016), SSD[2] (Liu et al. 2016), Faster R-CNN (Ren et al. 2016), Transformers and etc. These networks leverage machine learning algorithms and deep neural networks to achieve precise and rapid object detection.

The process of object detection can be broken down into two primary steps: identifying image regions likely to contain objects and then categorizing the objects within these regions independently. Previously, before the emergence of deep learning, this was done through the slider window approach. This approach entails applying an image window to different parts of the image and retaining only the predictions with higher probabilities. Presently, the field recognizes two main families of object detectors (Degola 2022).

Various network architectures form the backbone of object detectors by extracting image features crucial for detection tasks. Among them, ResNet[3] (He et al. 2015) is notable for its superior performance, primarily due to its use of skip connections, which add the output of a block back to its input. This design enhances gradient flow from deeper to shallower layers, utilizing early-extracted features during inference. As a result, ResNet has inspired the development of numerous deep networks that address previous training and accuracy challenges. ResNet models are typically identified by their layer count, with configurations ranging from 18 to over 1000 layers (Degola 2022).

Object detection models can be classified into **one-stage**, **two-stage**, **and hybrid** models, as illustrated in **Fig. 5**. Each category will be discussed in detail in the following sections.

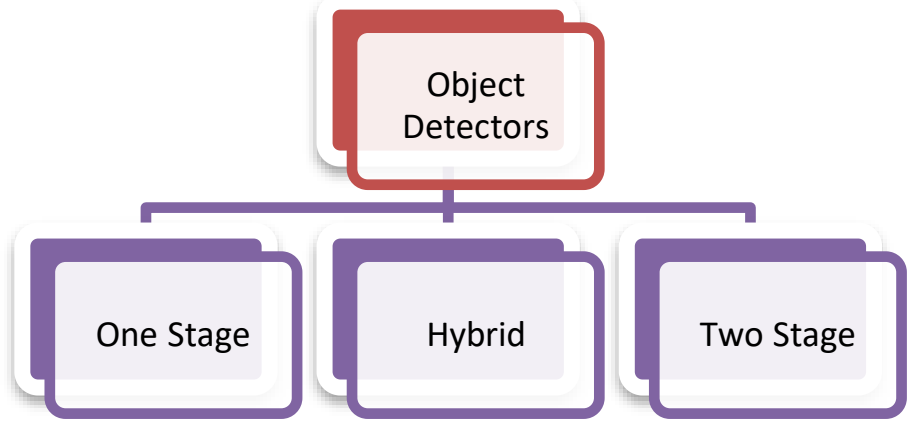

**Fig. 5.** The overview of Object Detectors

### Two-stage object detectors

[1] You Only Look Once
[2] Single Shot MultiBox Detector
[3] Residual Neural Network

Two-stage object detection models represent a class of object detection techniques that improve accuracy and precision by breaking down the detection process into two separate phases. In the first stage, potential object regions are identified, while in the second stage, these candidate regions are further refined and classified to produce the final detection results. This division allows for more focused processing, leading to enhanced detection performance compared to single-stage models (Ren et al. 2016).

**R-CNN:** Region-based Convolutional Neural Networks (R-CNN) (Girshick et al. 2014), was a groundbreaking model that significantly advanced the field of object detection. The model employs Selective Search to generate around 2000 region proposals from an input image, which are potential bounding boxes for object locations. These regions are then resized and passed through a Convolutional Neural Network (CNN), such as AlexNet (Krizhevsky et al. 2012), to extract feature vectors. Subsequently, these features are classified using Support Vector Machine (SVM) (Cristianini and Ricci 2008) classifiers trained to detect specific object classes, and a bounding box regressor refines the localization accuracy. R-CNN's modular design—separating region proposal, feature extraction, and classification—allows for independent enhancement of each component, leading to high accuracy in object detection benchmarks. However, it suffers from slow inference due to the computational cost of region proposal generation and CNN processing for each region, along with significant storage requirements for the features.

**Fast R-CNN:** Fast R-CNN (Girshick 2015), improved the efficiency of object detection by processing the entire image in a single forward pass through a CNN, generating a shared feature map, unlike its predecessor R-CNN, which processed each region proposal individually. A Region of Interest (RoI) pooling layer extracts fixed-size feature maps for each proposal, which are then classified via a softmax layer and refined using a bounding box regressor. This approach enhances speed and reduces memory usage by sharing computations, enabling end-to-end training. However, reliance on external region proposal methods like Selective Search remains a bottleneck for computational efficiency.

**RPN:** A Region Proposal Network (RPN) (Ren et al. 2016), integral to Faster R-CNN, streamlines object detection by efficiently generating high-quality region proposals through shared convolutional layers with the main detection network. Using a sliding 3x3 window over the feature map, it employs predefined anchor boxes of various scales and aspect ratios to predict objectness scores and refine bounding box coordinates. Non-Maximum Suppression (NMS) (Hosang et al. 2017) filters overlapping proposals, retaining the most promising ones. This shared-feature approach enhances detection speed and accuracy, and its adaptability has extended its use to frameworks like Mask R-CNN for instance segmentation and Feature Pyramid Networks (Lin et al. 2017) for multi-scale detection.

**Faster R-CNN:** Faster R-CNN (Ren et al. 2016), represents a major advancement in object detection by integrating region proposal generation directly into the network. Unlike R-CNN (Girshick et al. 2014) and Fast R-CNN (Girshick 2015), which rely on external methods for generating region proposals, Faster R-CNN uses a RPN (Ren et al. 2016) that shares convolutional features with the detection network, enabling faster and more efficient processing. The RPN predicts object bounds and objectness scores simultaneously, producing high-quality proposals. These proposals are then refined through a RoI pooling layer, and the resulting feature maps are used for object classification and bounding box regression. This design boosts speed and accuracy, supports end-to-end training, and proves effective for real-time applications across fields like autonomous driving, surveillance, and medical imaging.

**One-stage object detector**

One-stage detection models differ fundamentally from two-stage models by bypassing the region proposal phase and directly predicting object regions and classes within images. This streamlined approach significantly boosts prediction speed, making one-stage models particularly well-suited for real-time applications. However, this increased speed often comes with a trade-off in predictive accuracy. Pierre Sermanet et al. (Sermanet et al. 2014) introduced a key innovation in this field, replacing the final classification layers of a conventional convolutional neural network with a regression grid for each class, which outputs the coordinates of the objects' bounding boxes. Unlike two-stage detectors, which generate and refine region proposals before classification, one-stage detectors combine these steps into a single network pass, further enhancing their practicality for time-sensitive tasks (Degola 2022).

**YOLO:** The YOLO (You Only Look Once) (Redmon et al. 2016) model represents a significant advancement in deep learning for object detection and recognition in both images and videos. YOLO has gained prominence for its efficiency and speed in real-time object detection. This model revolutionizes object detection by framing it as a regression problem and segmenting images into a grid of cells. Each cell is responsible for detecting objects if the center of the object is within its boundaries, and it predicts the bounding box coordinates, confidence score, and class probability for those objects. YOLO utilizes a specialized loss function to simultaneously optimize the prediction of bounding box locations and object classification.

**SSD:** The Single Shot MultiBox Detector (SSD) (Liu et al. 2016) represent a robust and efficient approach to object detection by predicting class scores and refining bounding boxes for predefined default boxes using small convolutional filters applied to feature maps. Built on the VGG-16 (Simonyan and Zisserman 2015) architecture, SSD incorporates additional smaller convolutional layers to enhance its multi-scale feature representation, enabling accurate detection of objects of varying sizes at different network levels. By utilizing predefined reference boxes instead of segmenting images into grid cells, SSD achieves faster and more streamlined detection, making it suitable for rapid and accurate object detection across diverse scenarios.

**RetinaNet:** RetinaNet (Lin et al. 2018) represents a significant advancement over earlier one-stage object detection models, primarily through its incorporation of two key innovations: the Feature Pyramid Network (FPN) and focal loss. The integration of FPN into the architecture enhances the model's ability to detect objects at varying scales by creating a rich feature pyramid that improves the representation of objects of different sizes. Meanwhile, the adoption of focal loss addresses the challenge of class imbalance inherent in one-stage detectors by focusing more on hard-to-detect objects and reducing the impact of easy-to-classify background examples. These improvements collectively enable RetinaNet to achieve performance levels that are comparable to those of traditional two-stage detection models, which historically have had an edge in accuracy but with greater computational complexity.

**FPN:** The Feature Pyramid Network (FPN) (Lin et al. 2017) architecture is structured into multiple hierarchical levels, each corresponding to distinct stages within the network. At each stage, the network is composed of several convolutional layers of uniform size, with the layer sizes increasing by a factor of two at each subsequent stage. These stages are interconnected through a bottom-up and top-down pathway, with lateral connections facilitating communication between levels. This design enables the network to construct a detailed, multi-scale feature pyramid for every input image, which enhances its ability to detect objects of varying sizes. Since its introduction as a component in single-stage object detection models, FPNs have also been integrated into the framework of two-stage detection models. In this context, they contribute to improving the overall performance of these detectors by leveraging their capability to create rich and diverse feature representations across different scales.

**Hybrid object detector**

**Transformers** in object detection task can be used in both one-stage and two-stage detectors, depending on the specific architecture and implementation.

*One-Stage Detectors.*

A notable example of a one-stage detector is the DEtection TRansformer (Carion et al. 2020) model, which approaches object detection as a set prediction task. DETR streamlines the detection pipeline by directly predicting object bounding boxes and classes in an end-to-end manner, eliminating the need for intermediate region proposal steps. This method leverages the transformer's ability to model global relationships within the image, resulting in a simplified detection process with competitive performance. While DETR generally offers faster performance, it may exhibit slightly lower detection accuracy compared to other models.

**Detection-Transformer (DETR):** The DETR (Carion et al. 2020), streamlines object detection by eliminating the need for anchor boxes and traditional post-processing methods like Non-Maximal Suppression (NMS). Unlike conventional CNN-based detectors, DETR uses a unique architecture with three main components: a backbone network that incorporates positional encodings, an encoder, and a decoder with attention mechanisms. In this framework, the backbone extracts feature representations, combines them with positional encodings, and passes them to the encoder, where self-attention is applied to form multi-head attention outputs. These outputs are then processed in the decoder, which simultaneously decodes object queries and pairs predictions with ground-truth objects using a bipartite-matching algorithm, optimized by the Hungarian method and refined with Stochastic Gradient Descent (SGD). This architecture simplifies prediction by generating a fixed number of object boxes, set at 100 for the COCO (Lin et al. 2014) dataset. By employing parallel decoding and bipartite matching loss, DETR avoids the autoregressive decoding typical in prior models. Although DETR represents a major advance in simplifying detection processes, it faces challenges in training speed and small-object detection, which remain active areas for improvement (Carion et al. 2020; Shehzadi et al. 2023).

**Deformable-DETR:** The Deformable-DETR model (Zhu et al. 2021) improves the original DETR model by enhancing computational efficiency and feature resolution. Instead of applying uniform attention to all pixels, which is computationally intensive, it focuses on a smaller, relevant subset of pixels around a reference point, reducing complexity. It also employs a feature pyramid that integrates high and low-resolution features using relative positional embeddings, allowing the model to capture details at multiple scales. By replacing the traditional attention module with a multi-scale deformable attention mechanism, Deformable-DETR addresses issues of slow convergence and limited spatial resolution, making it more effective for large-scale applications and precise object detection tasks (Zhu et al. 2021; Shehzadi et al. 2023).

*Two-Stage Detectors*

Alternatively, transformer modules are integrated into two-stage detection frameworks to enhance feature representation and improve detection accuracy. These hybrid models, such as the Transformer-based Set Prediction with RCNN (TSP-RCNN) (Sun et al. 2021), first generate region proposals and then apply transformer-based modules for further refinement and classification. Although this approach often leads to higher accuracy, it typically demands more computational resources (Sun et al. 2021; Quan et al. 2024).

**TSP-RCNN:** TSP-RCNN (Sun et al. 2021) is a hybrid model that combines the Transformer architecture with the traditional Faster RCNN (Ren et al. 2016) to improve object detection accuracy and efficiency. It starts by generating region proposals using an RPN, processes these through a backbone network for feature extraction, and then uses a Transformer encoder-decoder to produce context-aware embeddings and final predictions. Unlike conventional detectors, TSP-RCNN outputs a set of predictions directly, reducing duplicate detections and enhancing accuracy. It employs Hungarian loss for object matching and standard RCNN loss for bounding box regression and classification, leading to faster convergence and better detection performance compared to the original DETR.

In the following, we provide a detailed comparison of various object detection models and their performance under domain adaptation settings. As shown in **Table 1**, we summarize key characteristics, including accuracy, speed, and robustness across different detection frameworks, allowing for a comprehensive understanding of how each model performs when adapted to new domains.

**Table 1.** Overview comparison between Detectors

| Category | Examples | Speed | Transferability | Domain Shifts | Strengths | Limitations | Complexity | Techniques |
|---|---|---|---|---|---|---|---|---|
| **One-Stage** | YOLO, SSD, RetinaNet | High | Transfer learning possible, but fixed feature extraction limits flexibility | Sensitive to appearance changes, lighting variations, and object scale differences | Suitable for real-time applications- Less computationally expensive | Slightly lower accuracy than two-stage models- Struggles with small objects | High: Requires extensive retraining, data augmentation, or fine-tuning for domain-specific adaptation | Domain Randomization, Data Augmentation, and Fine-Tuning |
| **Two-Stage** | Faster R-CNN, Cascade R-CNN | Moderate to Low | Modular design allows transfer of feature extractors (backbone networks) and proposal and refinement stages separately | More adaptable to domain changes due to separate steps | High accuracy for well-established tasks- Adaptable to domain changes | Requires high computational resources- Slower inference time | Moderate: Needs backbone fine-tuning but adaptable to different domains | Feature Alignment, Fine-Tuning Pre-Trained Backbones, Adversarial Domain Adaptation |
| **Hybrid** | DETR, Deformable DETR, TSP-RCNN | Moderate | Transformers capture global context, enabling flexibility in understanding object relations | Better at handling domain shifts, but convergence can be slow | Captures global context- Flexible in understanding object relations | Convergence can be slow- May struggle with small objects | Moderate to High: Self-supervised pre-training, Transformer-based Feature Alignment, and Fine-Tuning | Self-supervised Pre-training, Transformer-based Feature Alignment, and Fine-Tuning |

## 2.3 Domain Adaptation Method

Due to deep neural networks high accuracy and their alignment with cutting-edge advancements, they have become increasingly prevalent in various artificial intelligence and machine learning applications. Nevertheless, these models encounter significant challenges when dealing with domain shifts, as they often struggle to generalize across diverse data distributions beyond the source domain, resulting in suboptimal performance. Additionally, deep neural networks typically require substantial volumes of labeled data for effective training, which may not always be readily accessible due to time, resource, or feasibility constraints. This underscores the importance of transfer learning in deep neural networks. Unlike shallow learning, where domain adaptation is treated as a separate process, in deep learning, the focus is on embedding domain adaptation within the learning framework to acquire transferable representations (Singhal et al. 2023).

**Fig. 6** provides an overview of various deep domain adaptation techniques, building upon the classification proposed by Wang and Deng (Wang and Deng 2018). Despite this, the paper delves deeper into the techniques of deep domain adaptation specifically within the image domain. Our objective is to review and categorize deep domain adaptation approaches used for object detection, with a particular focus on feature-based methods.

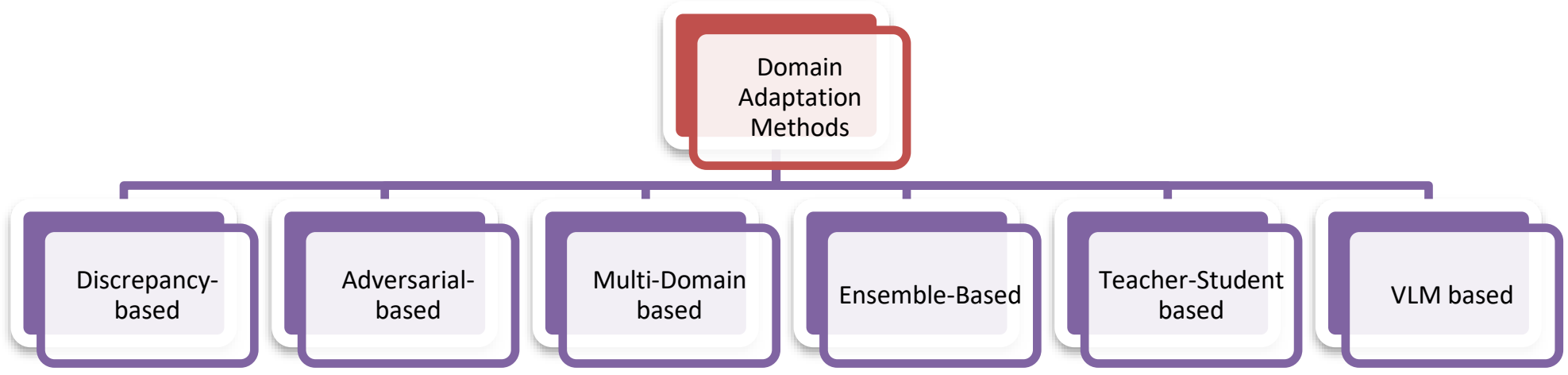

**Fig. 6.** The overview of Domain Adaptation Methods

**Discrepancy-based methods**

In the realm of discrepancy-based domain adaptation, a promising approach shifts the emphasis from conventional distribution alignment and feature transformation to measuring and minimizing the differences between source and target domains. The work by Mathelin et al. (de Mathelin et al. 2022) introduce a novel discrepancy-based active learning strategy for domain adaptation in object detection, shifting the focus from traditional distribution alignment to directly quantifying and minimizing domain gaps. Their approach leverages a discrepancy measure to link domain differences with target prediction accuracy, enabling selective labeling of impactful target samples to reduce generalization error. The framework is supported by theoretical bounds connecting the discrepancy measure to target risk, and a scalable algorithm incorporating regularization ensures compatibility with diverse loss functions. This method offers a robust alternative to adversarial training, enhancing model robustness and accuracy, and holds potential for integration with ensemble methods or other domain adaptation strategies.

Techniques like Maximum Mean Discrepancy (MMD) (Gretton et al. 2008) are employed to bridge gaps between different imaging modalities, yet they require extensive annotated data, which is often scarce in medical fields (Zhu 2024). Negative transfer can occur when irrelevant subdomains are included, leading to decreased model performance. Strategies to mitigate this include optimizing transfer weights to minimize feature distribution disparities (Xu et al. 2024b). Implementing discrepancy-based domain adaptation in real-world applications presents several key challenges that researchers must navigate. These challenges primarily revolve around data distribution alignment, theoretical underpinnings, and the risk of negative transfer. Ensuring maximum correlation between source and target domains is crucial, as mere alignment of marginal distributions may not suffice for effective classification (Lu and Huang 2024). The Step-wise Domain Adaptation DEtection TRansformer (SDA-DETR) (Zhang et al. 2024a) decomposes the adaptation process into three steps, progressively reducing domain discrepancies at both image and feature levels. This method constructs a target-like domain to facilitate knowledge transfer, enhancing model robustness against domain shifts. In **Fig. 7**, a general overview of this method is presented.

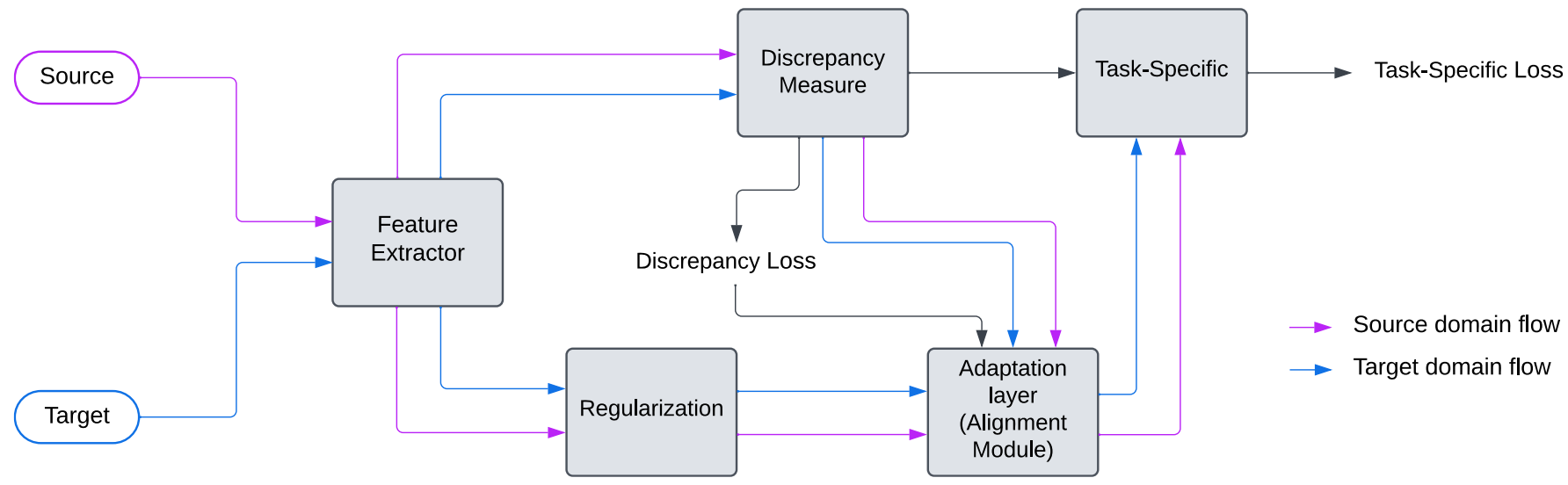

**Fig. 7.** A schematic representation of a discrepancy-based domain adaptation framework. This diagram depicts the general structure of a discrepancy-based domain adaptation framework. Key components include a Feature Extractor, Adaptation Layer, Discrepancy Measure, and Task-Specific Module. Discrepancy Loss guides the Task-Specific Loss while directly influencing the Adaptation Layer, enabling effective domain alignment. The flows for the source and target domains are distinctly marked, highlighting their interaction and feedback loops to minimize domain gaps and improve task performance. Best viewed in color.

**Adversarial-based methods**

The idea behind adversarial methods is that, alongside increasing domain ambiguity, they are trained robustly to understand domain separation. This concept is closely related to Generative Adversarial Networks (GANs) (Goodfellow et al. 2014), which consist of two components: a generator and a discriminator operating in a competitive environment. The generator's goal is to produce outputs that deceive the discriminator, while the discriminator, on the other hand, attempts to distinguish between real and synthetic images (Singhal et al. 2023).

In domain adaptation, the adapted idea is that the discriminator should be able to differentiate between the distributions of the source and target domains using domain-invariant features. Adversarial Discriminative Domain Adaptation (ADDA) (Tzeng et al. 2017) introduced a general framework for domain adaptation using adversarial models. The typical architecture for adversarial discriminative data matching employs a Siamese architecture with two data branches: one for source data and one for target data. This architecture is trained using a loss function (usually classification) or an adversarial or discrepancy-based loss function. In its simplest form, the adversarial generative architecture includes a generator that maps from one domain to another; then, the generated mapping and other mappings are connected to the adversarial discriminator architecture, and after being refined together, the generated mapping and other mappings adhere to the discriminator architecture (Singhal et al. 2023). YOLO-G (Wei et al. 2023) model incorporates feature alignment and unsupervised adversarial training through functional branches, complemented by a lightweight three-layer classifier to enhance detection accuracy with minimal computational overhead. Key methodologies include feature processing using FPN and CSP, H-divergence minimization through gradient reversal, and a combined loss function for detection and domain alignment. Feature uncertainty domain adaptation (FUDA) (Zhu et al. 2023b) introduced to improve detection precision in adverse weather conditions by aligning features based on local image blurriness. It also proposes an innovative adversarial learning strategy featuring an Instance-Level Uncertainty Alignment module that leverages feature channel entropy to guide alignment, addressing challenges in bounding box regression and classification caused by domain shifts. FDUA framework effectively utilizes unlabeled target domain data, demonstrating superior cross-domain detection performance under challenging weather conditions compared to traditional methods. D²-UDA (Zhu et al. 2023a) employs a disentangled discriminator to align feature distributions between source and target domains. The discriminator separates in-distribution features from outliers and incorporates a gated strategy to selectively process out-of-distribution samples, enhancing alignment through adversarial training. This mechanism is complemented by a Teacher-Student framework, where the teacher network generates pseudo labels refined iteratively via self-training, supported by parameter updates using Exponential Moving Average (EMA). Together, these adversarial components effectively address domain discrepancies, improving object detection performance in real-world scenarios. In **Fig. 8**, a general overview of this method is presented.

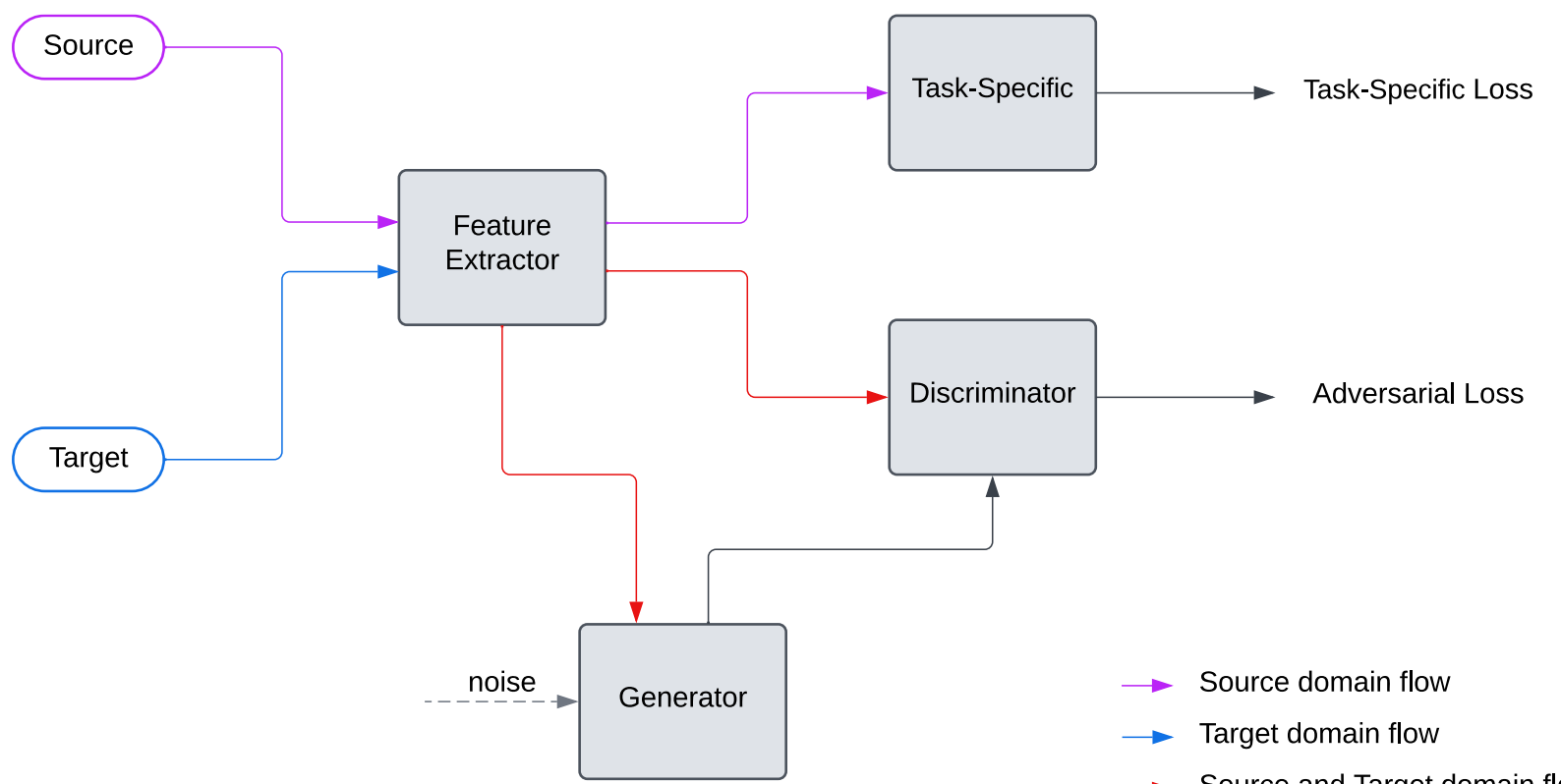

**Fig. 8.** A schematic representation of an adversarial-based domain adaptation framework. This diagram illustrates the general structure of adversarial-based domain adaptation methods. A feature extractor is employed to learn domain-invariant features from both source and target data. These features are utilized by the classifier for supervised learning on labeled source data and by the discriminator for adversarial training to align feature distributions across domains. A generator introduces noise to facilitate adversarial learning, aiding the discriminator in refining domain-invariant representations. Best viewed in color.

### Multi-Domain methods

Multidomain adaptation differs significantly from conventional domain adaptation models. In standard domain adaptation, there is typically one source domain and one target domain, where the model must adapt from the source domain to the target domain. However, in multi-domain adaptation, two scenarios may arise: multi-source adaptation and multi-target adaptation.

*Multi-source Adaptation.*

Multi-source domain adaptation (MSDA) for object detection improves detection models by utilizing multiple labeled source datasets and unlabeled target data to address domain distribution shifts. Unlike single-source adaptation, MSDA adapts from multiple sources to a single target domain, enhancing accuracy and robustness. Recent advancements focus on class-specific feature alignment to tackle challenges like noisy pseudo-labels and class imbalance, making MSDA a logical step for building more robust domain adaptation models.

Atif Belal et al. (Belal et al. 2024a) tackled the challenges of MSDA in object detection by proposing the Prototype-based Mean Teacher (PMT) approach. Traditional unsupervised domain adaptation (UDA) methods often fail in MSDA settings by aggregating multiple labeled source domains into a single entity, leading to reduced robustness and accuracy. Existing MSDA methods aim to learn domain-invariant and domain-specific features but face scalability issues due to high memory demands and overfitting risks as the number of source domains increases. PMT addresses these limitations by leveraging class prototypes to encode domain-specific information and aligning categories across domains through contrastive loss, achieving parameter efficiency without expanding the model with additional sources. Furthermore, the method avoids the limitations of class-agnostic alignment techniques, which neglect category-specific traits and reduce detection efficacy. Complementary approaches, such as attention-based class-conditioned alignment (ACIA) (Belal et al. 2024b), enhance MSDA by integrating attention mechanisms with adversarial domain classifiers to achieve domain-invariant, class-specific representations. Experimental results across diverse datasets validate the scalability, robustness, and superior performance of PMT and related methods, demonstrating their potential for advancing MSDA in real-world object detection.

Collaborative Learning (CL) (Cheng 2024) tackles the challenges of multi-source domain adaptation, an extension of single-source domain adaptation that involves addressing performance degradation caused by distribution discrepancies among multiple source domains. To overcome these challenges, the authors propose a novel method called CL, which preserves intra-domain knowledge across sources to improve the model's generalization capability. By leveraging collaborative strategies, the CL approach mitigates distribution differences and enhances adaptability in object detection tasks. Comprehensive experiments across diverse adaptation scenarios demonstrate the efficacy of CL in improving performance and robustness, addressing key limitations of existing methods. This work represents a significant contribution to the field of domain adaptation, offering insights into effective utilization of multiple source domains and paving the way for more versatile object detection systems.

The Dual-Level Alignment Network with Ensemble Learning (DANE) (Yang et al. 2024) addresses key challenges in MSDA by tackling both intradomain and interdomain shifts while leveraging knowledge embedded within and across source domains. To mitigate intradomain shifts, DANE introduces a clustering loss that utilizes a batchwise prediction similarity matrix (PSM) to create a discriminative and well-clustered feature space in the target domain. For interdomain shifts, it employs a dynamic weighting function that emphasizes samples with higher uncertainty, enhancing alignment between source and target distributions. Additionally, DANE leverages ensemble learning by constructing an ensemble source that integrates multiple source domains, enabling the extraction of domain-invariant knowledge. An ensemble-weighted decision integration strategy further combines knowledge from individual domains to optimize classification performance. Extensive experiments on five benchmark datasets demonstrate DANE's superior performance over existing methods, establishing it as the new state of the art. Its effectiveness is also validated in open-world scenarios, highlighting its robustness in diverse settings.

*Multi-target Adaptation.*

In this case, the number of target domains exceeds one. This means that the model must adapt from a single source domain to multiple distinct target domains. This concept is employed when a unified model is required to adapt to several different target domains, necessitating the ability to align with all these domains.

The MTDA-DTM (Nguyen-Meidine et al. 2022) addresses key challenges in multi-target domain adaptation (MTDA) for object detection, including catastrophic forgetting, computational complexity, domain shifts, and the absence of labeled data. By enabling incremental adaptation, it prevents the loss of knowledge about previously learned domains without requiring stored data. Unlike computationally intensive methods that rely on duplicated models, MTDA-DTM offers efficiency with minimal overhead. To tackle significant domain shifts, it incorporates a Domain Transfer Module, aligning source images with multiple target domains in a shared representation space. Moreover, the framework operates unsupervised, enabling adaptation to new domains without labeled data.

CoNMix (Kumar et al. 2023) combines Consistency with Nuclear-Norm Maximization and Mix-up knowledge distillation, as a solution. Addressing Single and Multi-Target Domain Adaptation under a source-free paradigm, the framework overcomes the challenge of unavailable labeled source data, often restricted by privacy concerns. By leveraging noisy target pseudo-labels, the approach enhances adaptation through label-preserving augmentations and pseudo-label refinement to mitigate noise. To achieve better generalization across multiple target domains, the framework employs Mix-up knowledge distillation, integrating multiple source-free Single Target Domain Adaptation models effectively.

**Teacher-Student methods**

In recent years, the deployment of deep neural networks (DNNs) on edge devices has posed significant challenges due to the large number of parameters and high computational costs associated with these models. To address these issues, model compression techniques such as knowledge distillation have gained prominence. Knowledge distillation involves transferring knowledge from a larger, more complex "teacher" model to a smaller, more efficient "student" model. This enables the student to achieve comparable performance while being optimized for real-time, resource-constrained applications. The Teacher-Student framework, widely discussed in this paper, provides a comprehensive review of various knowledge learning objectives, including knowledge distillation, expansion, adaptation, and multi-task learning. The architecture leverages a smaller student model that learns from the teacher model, reducing computational demands without sacrificing performance (Hu et al. 2022).

The Unified Multi-Granularity Alignment (MGA) (Zhang et al. 2024b) framework enhances object detection across domains through a two-stage process involving teacher-student training. It incorporates multi-granularity alignment across pixel, instance, and category levels to address domain adaptation challenges comprehensively. The Omni-Scale Gated Fusion (OSGF) module refines feature representations for objects of varying scales, while the Adaptive Exponential Moving Average (AEMA) strategy dynamically updates the teacher detector's parameters for generating high-quality pseudo labels. The training involves optimizing a teacher detector with source data and refining a student detector with both labeled source data and pseudo-labeled target data. Extensive ablation studies validate the contributions of components like OSGF, multi-granularity discriminators, and AEMA to the framework's performance improvements. In **Fig. 9**, a general overview of this method is presented.

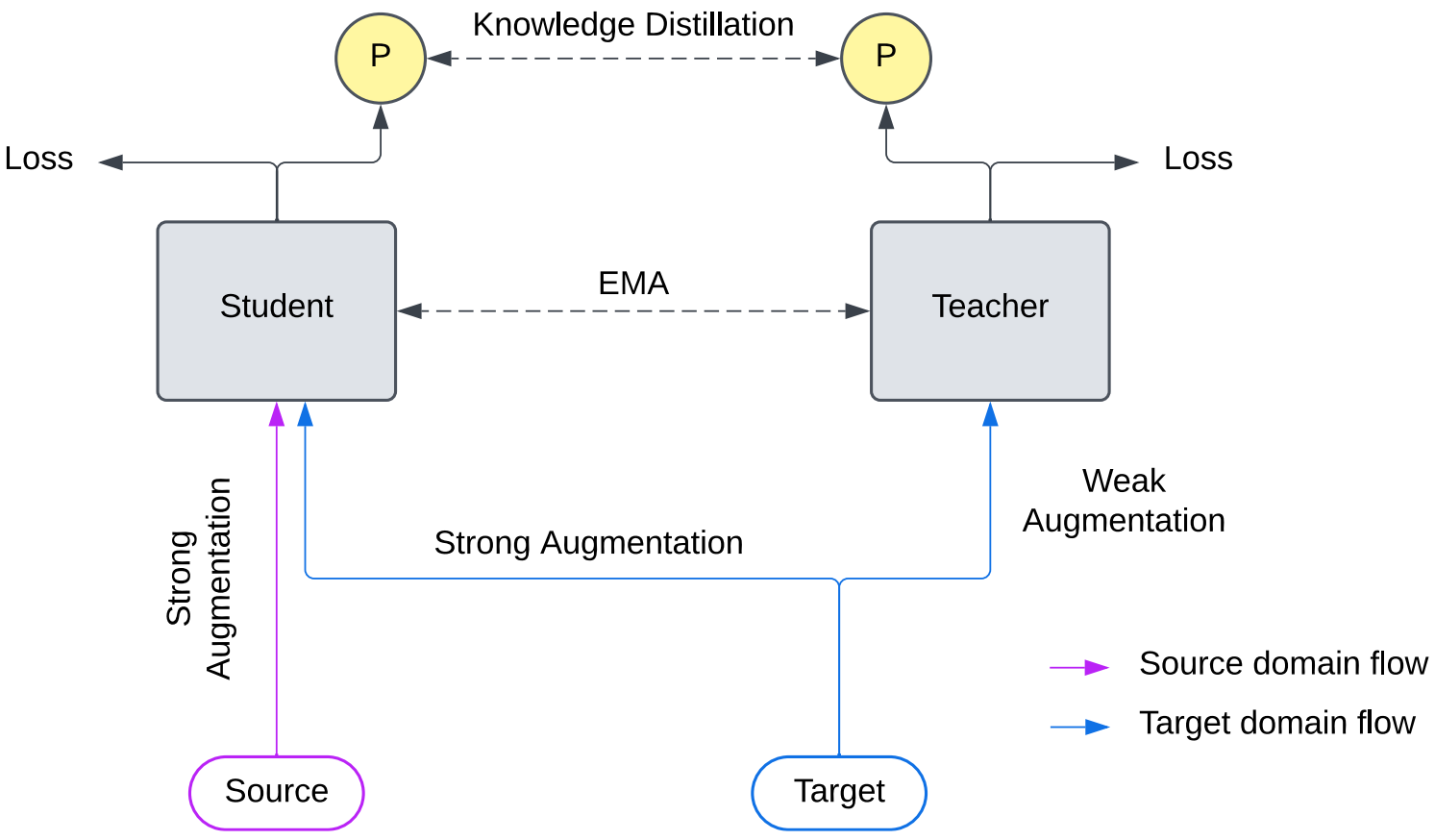

**Fig. 9.** A schematic representation of a Teacher-Student framework in domain adaptation. This diagram presents the general structure of the Teacher-Student framework in domain adaptation. The student network is trained using both source and target data, while the teacher network generates pseudo-labels for the unlabeled target data. These pseudo-labels (P) are iteratively refined through feedback between the teacher and student networks, enhancing the alignment of target domain features. The teacher's parameters are updated using mechanisms such as Exponential Moving Average (EMA), ensuring stability and robustness in cross-domain adaptation. Best viewed in color.

*Knowledge Distillation.*

Knowledge distillation aims to train a smaller, more efficient student model by leveraging the predictions generated by a larger, more complex teacher model. The goal of this process is to create a compact student model that retains performance levels similar to those of the teacher model, achieving efficiency without a significant loss in accuracy (Hu et al. 2022).

The knowledge distillation process involves a two-phase training approach. Initially, the teacher model is trained on a large dataset to learn intricate patterns and make accurate predictions. Once the teacher is trained, it generates outputs that include both hard labels (the actual class labels) and soft labels (probability distributions over the classes). These soft labels provide additional insights into the relationships between classes, which are crucial for the student model's learning process. In the second phase, the student model is trained using the outputs from the teacher. The student is tasked with two objectives: it must predict the ground truth labels accurately and also match the softened label distributions produced by the teacher. This dual training mechanism allows the student to learn not only the correct answers but also the underlying decision-making process of the teacher, thereby achieving performance that closely resembles that of the teacher model while being computationally efficient (Hu et al. 2022).

Liang Yao et al. (Yao et al. 2025) presents a progressive knowledge distillation approach aimed at reducing the feature gap between teacher and student models in UAV-based object detection. By progressively transferring knowledge, the method ensures that the student model efficiently assimilates the intricate patterns learned by the teacher model. The approach is particularly effective in handling the domain shift encountered in UAV imagery, leading to improved detection accuracy and robustness in diverse environmental conditions. Minhee Cho et al. (Cho et al. 2024) proposes a collaborative learning framework that enhances unsupervised domain adaptation through a unique teacher-student interaction. The method updates the teacher model's non-salient parameters using feedback from the student model, fostering a mutual learning process. This collaborative approach not only improves the performance of both models but also ensures better adaptation to the target domain. The results indicate substantial gains in object detection accuracy, highlighting the potential of collaborative learning in domain adaptation. Haozhao Wang et al. (Wang et al. 2023) introduces a domain-aware federated knowledge distillation method that treats local data in each client as a specific domain. The proposed approach optimizes the ensemble of soft predictions from diverse models by incorporating domain-specific knowledge into the distillation process. This method enhances the generalization ability of the student model across different domains, making it particularly suitable for federated learning environments where data privacy and heterogeneity are critical concerns.

*Knowledge Expansion.*

The concept of knowledge expansion involves using a teacher model's predictions as pseudo-labels for training a student model, potentially outperforming the teacher through iterative self-training. This section explores strategies like data augmentation and curriculum learning to enhance object detection tasks, ensuring the generalization of student models across various domains (Hu et al. 2022).

The Contrastive Mean Teacher (CMT) (Cao et al. 2023) framework addresses the challenges of domain discrepancies and noisy pseudo-labels in unsupervised domain adaptation (UDA) for object detection. By combining mean-teacher self-training with object-level contrastive learning, CMT enhances feature adaptation and extracts robust representations even when pseudo-labels are imperfect. This innovative integration of self-training and contrastive learning establishes a new approach for domain adaptation in object detection, achieving consistent improvements over existing methods as demonstrated by extensive evaluations across diverse datasets and adaptation tasks.

*Knowledge Adaptation.*

Knowledge adaptation facilitates the transfer of knowledge between domains while addressing challenges like catastrophic forgetting. The process often begins with training a teacher model on a well-defined source domain. This teacher then guides a student model, focusing on areas where the student struggles most, ensuring effective learning and generalization across new tasks (Hu et al. 2022).

knowledge adaptation can also involve advanced techniques such as adversarial training. For instance, Judy Hoffman et al. in their work (Hoffman et al. 2017) use CycleGAN (Zhu et al. 2017) to align source and target domain images, enabling adversarial training where the discriminator functions as the teacher, guiding the student to learn from both domains effectively (Hu et al. 2022). Advanced methods extend knowledge adaptation to scenarios like source-free domain adaptation. Vibashan VS et al. (Vs et al. 2022) proposed leveraging an Instance Relation Graph (IRG) to model object relationships in the target domain. By using a contrastive loss guided by the IRG and student-teacher distillation to handle noisy pseudo-labels, their approach improves generalization even without source data. Further innovations include techniques like Masked Image Consistency (MIC) (Hoyer et al. 2023), which addresses annotation challenges in visual recognition. MIC employs random patch masking, a consistency loss function, and pseudo-labeling with an EMA teacher to infer spatial relationships and enhance performance across tasks such as semantic segmentation, image classification, and object detection.

*Multi-Task Learning.*

Multi-task learning enables a student model to learn from multiple teacher models across different tasks. The section emphasizes the alignment of predictions with ground truth and highlights the benefits of cross-task knowledge distillation to improve the generalized representation of knowledge (Hu et al. 2022).

**Ensemble methods**

Ensemble learning has emerged as a powerful approach for addressing domain adaptation challenges and managing complex data characteristics by leveraging the collective strengths of multiple models. This method effectively mitigates domain shift through strategies such as diversity in model training, weighted voting based on target-domain performance, and integration with transfer learning to fine-tune models for new domains. Additionally, ensemble methods utilize semi-supervised learning with unlabeled target data, adaptive resampling techniques to balance training distributions, and domain-specific feature selection to enhance generalization. They also employ feedback mechanisms to iteratively refine performance on target data and adaptively prioritize difficult-to-predict instances. Furthermore, ensemble approaches handle noise, outliers, and imbalanced datasets by aggregating predictions across models, emphasizing challenging cases, and incorporating feature diversity to capture complex relationships. These strategies collectively enable ensemble learning to deliver robust, reliable, and adaptable performance in scenarios where traditional models struggle with domain-specific variations and real-world data intricacies (Dong et al. 2020). Ensemble techniques help address complex domain shifts and reduce gap between source and target domains by merging predictions from multiple models. They offer unique mechanisms to improve model transferability across domains and reduce the need for labeled target data.

Pseudo-labeling is a prominent ensemble-based method for UDA that iteratively generates labels for unlabeled target data, allowing the model to adapt its predictions over time. While pseudo-labeling has shown promise in reducing the need for human-labeled target data, it faces challenges associated with noisy labels, which can degrade performance if incorrectly learned. To address this, methods like DebiasPL (Wang et al. 2022b) incorporate label filtering techniques to eliminate low-confidence pseudo-labels, thereby improving label accuracy and stabilizing model adaptation in each iteration.

In Domain Adaptation tasks, ensembles extend their utility. For instance, Synthetic Labeling Aggregation assigns pseudo-labels to target domain samples when multiple models agree with high confidence. Multi-source ensemble models, such as MS3D++ (Tsai et al. 2023), further enhance pseudo-labeling by aggregating predictions from several "expert" models, each trained on different source domains. By combining outputs from these models, MS3D++ effectively captures domain-invariant features and mitigates the effects of negative transfer, where irrelevant information from source domains might harm target performance. Additionally, domain-aware approaches, such as DA-Pro (Li et al. 2023), refine pseudo-labeling by introducing domain-specific prompts. These prompts allow the model to dynamically adjust to target domain features based on unique domain attributes, which is particularly advantageous in environments with varying conditions, such as autonomous driving.

**Vision Language Models Methods**

Vision-Language Models (VLMs) combine visual and linguistic information, enabling them to generalize across tasks and domains. Their ability to process multimodal data makes them especially effective for addressing domain adaptation challenges in object detection, where the source and target data distributions differ significantly. Below are key domain adaptation methods leveraging VLMs in object detection:

VLDadaptor (Ke et al. 2024) utilizes VLMs to distill knowledge into object detectors, enhancing their ability to generalize across domains. By aligning visual features with corresponding textual descriptions, models can better handle domain shifts. Designed for Incremental Vision-Language Object Detection, ZiRa (Deng et al. 2024) introduces a zero-interference loss and reparameterization strategies. This method allows models to adapt to new domains incrementally while preserving their zero-shot generalization capabilities, ensuring they remain effective across diverse tasks. DA-Pro (Li et al. 2023) employs learnable domain-adaptive prompts to generate dynamic detection heads tailored for each domain. By incorporating domain-invariant and domain-specific tokens along with textual descriptions, this method captures shared and unique domain knowledge, facilitating effective domain adaptation. CDDMSL (Malakouti and Kovashka 2023) leverages vision-language pre-training to align features through the language space. By maximizing agreement between descriptions of images with different domain-specific characteristics, models achieve better generalization across domains, addressing challenges in semi-supervised domain generalization for object detection.

Bridging the domain gap between curated datasets and real-world applications remains a significant challenge in UDA. Traditional ImageNet-pretrained CNNs perform well on small benchmarks but struggle with scalability and overfitting on larger datasets. Vision Transformers (ViTs) and VLMs, such as CLIP (Radford et al. 2021) and GLIP (Li et al. 2022), offer improved feature alignment via language supervision but face similar limitations. Zhengfeng Lai et.al introduces parameter-efficient VLM adaptations (Lai et al. 2024), including Prompt Task-dependent Tuning and Visual Feature Refinement, to enhance semantic representation and domain disentanglement. Extensive experiments on classification and detection tasks demonstrate substantial performance gains, with domain-aware pseudo-labeling emerging as a particularly effective strategy for aligning target domains while preserving pre-trained knowledge.

Traditional approaches often struggle with biases tied to the source domain, making them less effective for real-world applications. VLMs tackle this issue by leveraging large-scale image-text pretraining, offering more general and adaptable representations. However, the reliance on frozen encoders limits their flexibility for specific tasks or domains. The Domain-Aware Adapter (DA-Ada) (Li et al. 2024) provides a solution by combining domain-invariant and domain-specific adapters to align features and recover important target domain knowledge. It also introduces a Visual-guided Textual Adapter to improve detection by bridging visual and textual features. Validated on cross-weather and Sim-to-Real benchmarks, DA-Ada achieves state-of-the-art results through a balanced optimization framework combining adversarial, cross-entropy, and discrepancy losses, showcasing VLMs' potential in advancing robust domain adaptation.

Knowledge Graph Distillation (KGD) (Jiang et al. 2024) addresses the domain adaptability challenges of large-vocabulary object detectors (LVDs) by tackling discrepancies in data distribution and object vocabulary. Through CLIP-based encoding, KGD constructs a knowledge graph (KG) that captures semantic relationships within downstream data and integrates it into LVDs for enhanced cross-domain classification. Its dual extraction of visual and textual KGs provides complementary insights, improving both object localization and recognition. Extensive experiments on benchmark datasets demonstrate KGD's effectiveness, significantly outperforming state-of-the-art methods and advancing domain adaptation in large-vocabulary object detection.

## 2.4 Feature-Based Adaptation Approaches

The core concept here revolves around creating a shared feature embedding or representation by minimizing differences in data distribution. These methods strive to maintain the intrinsic characteristics of the input data while bridging distribution gaps (Singhal et al. 2023). Feature-based adaptation approaches are central to this effort, emphasizing the enhancement, alignment, and transformation of features to achieve better performance across varying scenarios. This article explores three key categories within this domain: Feature Augmentation/Reconstruction, Feature Alignment, and Feature Transformation.

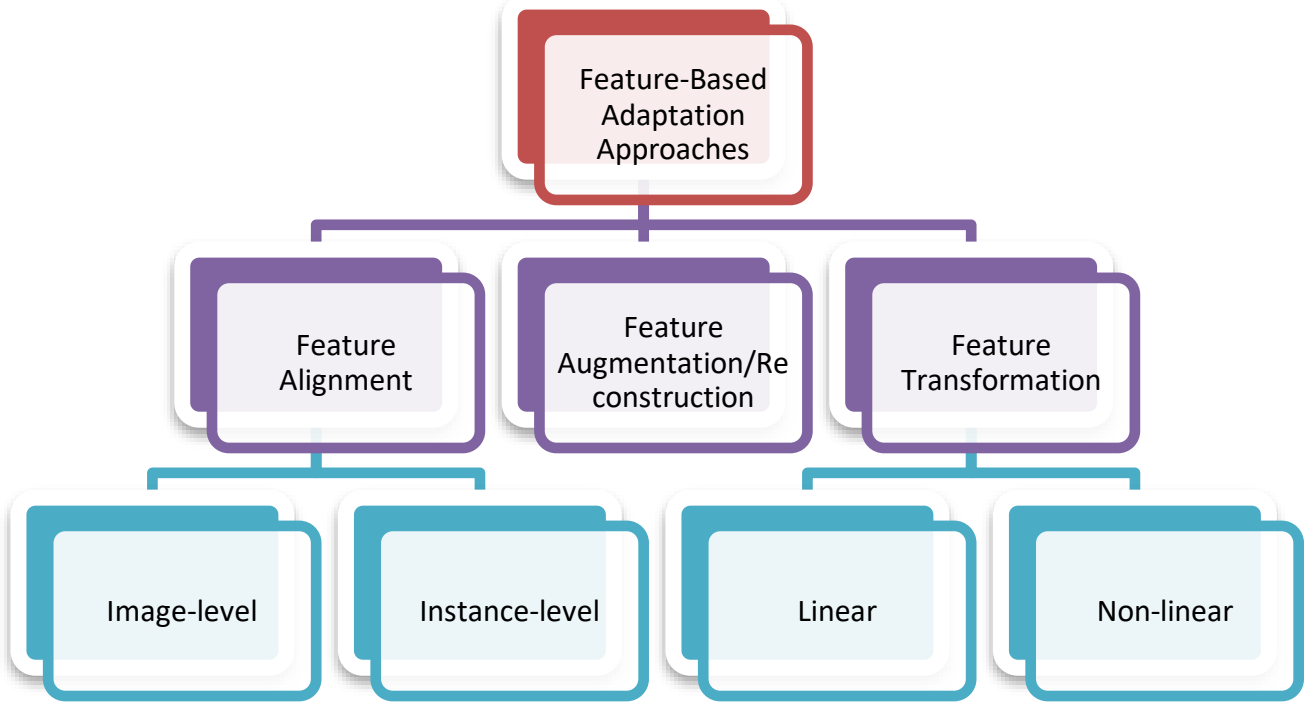

**Fig. 10.** Categorization of Feature-Based Adaptation Approaches

**Feature Alignment**

Feature alignment in domain adaptation is a critical technique aimed at reducing the discrepancy between the feature distributions of the source and target domains. Feature alignment refers to the process of transforming and aligning the feature representations from a source domain to a target domain. The primary goal is to create a common feature space where both domains exhibit similar characteristics, thereby enhancing the model's ability to generalize across different data distributions. The DA-DETR (Zhang et al. 2023) introduces a CNN-Transformer Blender (CTBlender) for the fusion and alignment of features from labeled source data and unlabeled target data, improving detection performance across domains. Key innovations include a shuffling mechanism for diverse feature representations and the Scale Aggregation Fusion (SAF) technique, which integrates multi-scale features with scale-specific weights. Extensive experiments show that DA-DETR outperforms existing domain adaptive object detection methods, especially in challenging scenarios like varying weather and diverse scenes. This work highlights the effectiveness of advanced feature fusion strategies in addressing domain gaps and enhancing model robustness. Semi-Supervised Domain Adaptive YOLO for Cross-Domain Object Detection (SSDA-YOLO) (Zhou et al. 2023) method is based on the advanced one-stage detector YOLOv5. It contains four main components: the Mean Teacher model with a knowledge distillation framework for guiding robust student network updating, the pseudo cross-generated training images for alleviating image-level domain differences, the updated distillation loss for remedying cross-domain discrepancy, and the novel consistency loss for further redressing cross-domain objectness bias.

By combining semi-supervised learning with knowledge distillation, the approach utilizes labeled source data, a small labeled target subset, and a large unlabeled target dataset to reduce domain gaps efficiently. Key innovations include a teacher-student model for feature alignment and global scene style translation for domain invariance. The method demonstrates superior performance in cross-domain object detection, showcasing the potential of modern single-stage architectures.

Feature alignment can be achieved at different levels, image-level and instance level.

*Image-Level.*

Image-level alignment focuses on aligning entire images from different domains to ensure that their overall feature representations are similar. This process often involves transforming or adjusting images to match the characteristics of another set of images, thereby minimizing the differences in their feature distributions. This can be done using style transfer methods, where the style of the source images is transformed to match the target domain (Nakamura et al.). Another approach is to use adversarial training to align the feature distributions of the source and target domains at the image level.

Jinlong Li et al. introduced CAST-YOLO (Liu et al. 2023b), that specifically designed to mitigate performance degradation due to domain shifts, particularly in foggy weather conditions. By integrating a cross-attention strategy transformer, the proposed model achieves effective feature alignment between source and target domains, resulting in improved detection accuracy. The framework also incorporates a Mean Teacher model for knowledge distillation and a convolutional block attention module to enhance feature focus and suppress noise. Experimental results demonstrate that CAST-YOLO outperforms existing methods, particularly in handling occlusions and varying scales, with superior performance on the Foggy Cityscapes dataset. These findings suggest that CAST-YOLO offers a robust solution for adaptive object detection in adverse weather, with potential applications extending to other challenging conditions.

*Instance-Level.*

Instance-level alignment involves aligning features of individual objects or instances within the images. This can be achieved through techniques such as instance normalization, where the features of each instance are normalized to reduce domain-specific variations (Li et al. 2024b). Additionally, instance-level adversarial training can be used to ensure that the features of objects from different domains are indistinguishable.

ALDI[4] (Kay et al. 2024), designed to address the challenges of DAOD by integrating key elements of feature alignment and self-training into a cohesive system. ALDI employs a student-teacher model. Building on ALDI, the enhanced ALDI++ method incorporates robust pretraining strategies and soft distillation techniques, significantly improving the quality of pseudo-labels and achieving state-of-the-art performance across multiple benchmarks. The Unified Multi-Granularity Alignment (MGA) (Zhang et al. 2024b) framework aligns features across pixel, instance, and category levels, addressing the interdependencies among these granularities.

There is the difficulty in achieving accurate object detection when applying the DETR framework across different domains, a common issue due to shifts in data distribution that can reduce model performance. Mean Teacher DETR with Masked Feature Alignment (MTM) (Weng and Yuan 2024) bypasses anchor-based methods in favor of object queries, has demonstrated strong results, it struggles with domain adaptation. MTM building on the two-stage Deformable DETR to improve DETR's adaptability and efficiency. MTM begins with a CycleGAN (Zhu et al. 2017) pretraining phase to simulate the target domain, followed by a self-training phase that leverages pseudo-labels and introduces techniques like Object Queries Knowledge Transfer (OQKT) and masked feature alignment. These enhancements significantly bolster the model's robustness and prevent stagnation in training, resulting in a marked improvement in detection accuracy across varying domains.

**Feature Augmentation/Reconstruction**

Feature augmentation and reconstruction techniques aim to enhance the robustness and generalizability of features extracted from images. These methods often involve generating additional data or modifying existing features to better represent the target domain.

Feature Augmentation involves creating synthetic variations of the existing data to increase the diversity of the training set. Techniques such as data augmentation (e.g., rotations, translations, and scaling) can be applied at the feature level to simulate different conditions and improve the model’s adaptability.

Feature Reconstruction focuses on reconstructing features to reduce domain discrepancies. Autoencoders and generative adversarial networks[5] are commonly used to learn a compact representation of the features, which can then be reconstructed to match the target domain's characteristics (Zhu et al. 2024). JFDI (Qiao et al. 2024) focuses on learning target-specific features through a dual-path architecture, enhancing adaptability by integrating source and target domain characteristics. It employs a hierarchical pseudo-label fusion module to improve the reliability of learned features. Masked Retraining Teacher-Student Framework (MRT) (Zhao et al. 2023), addresses the issue of domain shift by integrating a teacher-student paradigm with a strategy of masked retraining. Within this framework, a student model is developed using the target domain while a pre-trained teacher model, which has been trained on the source domain, offers support through feature maps and pseudo-labels. In order to alleviate the adverse impacts of domain shift, MRT integrates a masked retraining mechanism, which systematically masks certain regions of the teacher's feature maps throughout the training process. This approach incentivizes the student model to depend more on its intrinsic learning processes, thereby augmenting its adaptability to the target domain.

[4] Align and Distill
[5] GANs

Domain Disentanglement Faster-RCNN (DDF) (Liu et al. 2023a) introduces a robust feature disentanglement strategy to remove source-specific information through two key components: the Global Triplet Disentanglement (GTD) module, which adapts features at a global level, and the Instance Similarity Disentanglement (ISD) module, which focuses on local feature consistency. This dual-module design enables DDF to outperform leading UDA methods in object detection across four benchmark datasets, enhancing feature reliability and accuracy by minimizing domain-specific influences. This technique addresses the limitations of adversarial training by ensuring that extracted features are more domain-invariant. The reconstructed feature alignment network (RFA-Net) (Zhu et al. 2022) is designed to enhance unsupervised cross-domain object detection, particularly in remote sensing imagery, through modules dedicated to data augmentation, sparse feature reconstruction, and pseudo-label generation. RFA-Net employs a sequential data augmentation module at the data level to yield substantial improvements with unlabeled data, a sparse feature reconstruction module at the feature level to strengthen instance features for effective alignment, and a pseudo-label generation module at the label level to supervise the unlabeled target domain. This approach addresses critical challenges in instance-level feature alignment and noise reduction, which are essential for achieving accurate detection.

**Feature Transformation**

Feature transformation methods focus on transforming the features extracted from the source domain to better match those of the target domain. These transformations can be linear or non-linear and are designed to reduce the domain gap.

*Linear Transformations*

Linear transformations, such as PCA[6] (Shlens 2014) and LDA[7] (Xanthopoulos et al. 2013), can be used to project the features into a new space where the domain differences are minimized (Zhu et al. 2024). These methods are computationally efficient and can be easily integrated into existing pipelines. Linear feature transformations offer a straightforward approach to adapting features across domains, relying on simple mathematical operations like scaling, rotation, and translation. This simplicity promotes efficient computation and smooth integration into various machine learning frameworks, making linear transformations particularly useful in applications with real-time requirements due to their low computational demand. They are easy to implement and offer interpretability, as the transformation maintains a clear input-output relationship. However, their limitations lie in their restricted flexibility and assumption of linearity, which can be insufficient for complex domain shifts, where intricate relationships between source and target domains are better captured by nonlinear methods.

While effective for many scenarios, linear mappings may struggle with complex relationships between source and target distributions. In cases where non-linear transformations are necessary, more advanced techniques like multi-kernel learning may be employed instead (Han and Ling 2022).

In cross-domain object detection, linear transformations can be used to map features from a source dataset (such as urban scenes) to a target dataset (like rural scenes), enhancing detection accuracy when models trained on one domain are applied to another. These transformations employed in various applications, including EEG-based emotion recognition and hyperspectral image classification, where they help align marginal and conditional distributions effectively without requiring labeled samples from the target domain (Chai et al. 2017).

[6] Principal component analysis

[7] Linear discriminant analysis

*Non-Linear Transformations.*

Non-linear transformations, such as those achieved through deep neural networks, can capture more complex relationships between the source and target domains. Techniques like DANN[8] and GRL[9] are commonly used to learn domain-invariant features through non-linear transformations (Zhu et al. 2024). Nonlinear methods employ advanced mathematical functions, such as neural networks and kernel techniques, to create complex mappings that bridge source and target domains. These methods provide greater flexibility, allowing them to capture intricate, nonlinear relationships within data, which makes them especially useful in cases of significant domain shifts or complex feature structures. Nonlinear transformations often lead to improved domain alignment and enhanced robustness against variations like lighting, viewpoint, or occlusions. However, their complexity comes with higher computational costs and longer training times. Additionally, their increased capacity can raise the risk of overfitting, particularly when labeled data in the target domain is limited.

Non-linear transformations can effectively manage more complex types of domain shifts than linear transformations. For example, they can adapt to changes in feature distributions that are not merely shifts or rotations but involve more intricate alterations in data representation. Some approaches focus on learning asymmetric non-linear transformations, which allow for different mappings from the source to the target domain. This flexibility is beneficial when the two domains have different feature spaces or dimensions (Kulis et al. 2011; Kouw et al.).

The Localization Regression Alignment (LRA) (Piao et al. 2024) reframes localization regression as a classification task, leveraging adversarial learning to align features and enhance localization across domains. This approach directly addresses domain shifts, which often reduce the performance of object detection models trained on labeled source data but applied to unlabeled target data. Recognizing the distinct nature of features used for classification versus localization, the authors propose a comprehensive feature alignment that includes both types, moving beyond traditional methods focused on classification alignment alone. By converting regression into classification through discretization, LRA enables adversarial learning for more robust feature alignment. Additionally, the novel bin-wise alignment (BA) strategy further improves cross-domain feature alignment, enhancing object detection accuracy in target domains. Extensive experiments validate these methods, demonstrating that LRA achieves state-of-the-art performance and highlighting the importance of effective feature transformation in unsupervised domain-adaptive object detection.

BlenDA (Huang et al. 2024) introduces as a novel regularization approach for domain adaptation in object detection by generating pseudo samples from intermediate domains through feature transformation. Using a pre-trained text-to-image diffusion model, BlenDA blends source images with their translated counterparts, guided by target domain text labels. This blending process creates soft domain labels, enabling the model to learn from these pseudo samples and better bridge the source-target domain gap. As a result, BlenDA enhances the model's adaptability to target domain variations, leading to notable improvements in object detection performance.

---

[8] Domain adversarial neural networks

[9] Gradient reversal layers

# 3 Validation metrics

Evaluating the performance of object detection algorithms is a cornerstone of machine learning research, especially in the domain of computer vision. Unlike traditional image classification tasks, object detection is inherently more complex due to the need to accurately identify both the class and location of multiple objects within an image. To ensure fair and meaningful evaluation, well-defined metrics are essential. This section explores the foundational metrics of *precision*, *recall*, *Intersection over Union (IoU)*, and *mean Average Precision (mAP)*, which collectively provide a comprehensive assessment of object detection models.

**Precision and recall metrics**

In this part, we shall delve into the fundamental principles and prevalent metrics used to assess the efficacy of algorithms designed for object detection. The process of evaluating algorithms is an essential component of machine learning endeavors and can be executed through a variety of methodologies, contingent upon the specific task and objectives at hand. The task of object detection poses a higher degree of complexity compared to the classification of conventional images due to the potential presence of a variable number of objects within an image, necessitating the detection model to accurately forecast both their class and position. Consequently, it becomes imperative to establish and delineate precise evaluation criteria to facilitate this process. When considering the precision and recall criteria, it is paramount to elucidate on certain foundational concepts within the context of object detection. The TP[10] Value signifies a scenario where a finite box is correctly attributed to an object that is verifiably present in the image. Conversely, the FP[11] Value denotes a false detection event, wherein an object is erroneously identified within the finite box, either in part or whole. Furthermore, the FN[12] Value indicates instances where the object existing in the image remains undetected by the object detection model. The notion of a TN[13] loses relevance within the domain of object detection since the failure to detect an object that is absent in the image holds no significance. Essentially, this implies that the model has accurately discerned the background as not pertaining to any class. Precision serves as a metric of accuracy, gauging the precision of the model's predictions by furnishing a percentage or fraction of correct prognostications. In essence, this metric quantifies the extent to which the model's predictions are devoid of inaccuracies among all the forecasts made. The calculation of precision is contingent upon the following formula:

$$Precision = \frac{TP}{TP+FP} \quad (1)$$

Recall, on the other hand, acts as a yardstick to evaluate the recognition prowess of the model, providing a percentage or fraction of objects that have been appropriately detected. Essentially, this criterion delineates the proportion of correctly identified objects out of the total objects present in the image. The computation of recall is based on the following equation:

$$Recall = \frac{TP}{TP+FN} \qquad (2)$$

[10] True Positive
[11] False Positive
[12] False Negative
[13] True Negative

**IoU[14] metric**

For each anticipated finite box, this standard evaluates its subscript in relation to the actual finite box to generate a forecast categorized as either true positive or false positive. In the case of a finite pair of boxes, the Intersection over Union (IoU) criterion is determined in the subsequent manner:

$$IoU(A,B) = \frac{A \cap B}{A \cup B} \qquad (3)$$

IoU signifies the intersection area between the bounded boxes, divided by the union area of both constrained boxes. A higher IoU value indicates a greater level of conformity and precision in object detection. The calculation involves assessing the proportion of the image that is shared by both bounded boxes, as well as the combined area of the two restricted boxes. A high IoU value suggests a heightened level of agreement and accuracy in the identification of objects within the imagery.

**MAP metric**

The mean average precision[15] is an essential and widely used metric for evaluating the performance of deep learning models in object recognition and imaging tasks. This metric is highly valued in various fields such as computer vision, image processing, and virtual reality due to its effectiveness.

A key aspect of mAP is its ability to assess how accurately a model can detect objects by calculating the average precision across all categories and objects in the dataset. This comprehensive approach provides a holistic view of the model's performance.

The computation of mAP involves several sub-criteria, including the confusion matrix, intersection over union (IoU), recall, and accuracy. These criteria deeply examine the model's performance in object detection and precise localization. In the mAP calculation process, the model generates prediction scores for each image, which are then translated into different class labels. Subsequently, a confusion matrix is constructed, including TP, FP, TN, and FN.

From the confusion matrix, accuracy and recall metrics for each class are derived. The area under the accuracy-recall curve is then calculated for individual classes, culminating in the computation of mAP through a weighted average of these accuracy values across all classes. This weighting scheme ensures that more significant classes carry more weight in the evaluation.

The average accuracy within a set denotes the mean accuracy score for each member of that set, calculated according to a specific formula. mAP stands as a comprehensive benchmark for evaluating object detection algorithms, highlighting the model's ability to detect and localize objects accurately while addressing false positives and false negatives. The mAP for a set is the

[14] Intersection over Union

[15] mAP

average of the average precision scores for each member of that set. This metric is calculated using the following formula:

$$Precision = \frac{TP}{TP+FP} \quad (4)$$

Moreover, mAP outperforms other evaluation metrics by considering all relevant aspects of the evaluation process without focusing only on a limited number of high-ranking items. It evaluates all objects in an image, providing detailed insights into the model's performance across various classes.

Furthermore, mAP enables researchers to compare algorithms, make informed decisions regarding performance optimization, and fine-tune parameters. By conducting multiple evaluations across diverse datasets, a more precise assessment of the model's performance and the effects of any modifications can be obtained. Within the mAP evaluation criteria, thresholds such as 0.5 and 0.75 serve as indicators of success in data retrieval. mAP 50 evaluates the model's average accuracy at a 50% overlap threshold (IoU), emphasizing accurate object detection and classification.

Conversely, mAP 50-95 assesses the model's average accuracy across a range of overlap thresholds from 50% to 95% in 5% increments. This criterion provides a comprehensive evaluation of the model's performance by considering both high overlap scenarios (high accuracy) and low overlap instances (high recall).

# 4 Datasets

In this section, we examine the datasets commonly used in domain adaptation for object detection tasks. Domain adaptation typically requires two datasets for training: a source domain and a target domain. These datasets often have similar or completely different distributions, which poses challenges by reducing model accuracy and performance. The primary goal of domain adaptation is to align these distributions, thereby improving the accuracy and robustness of deep learning models. Among the widely used dataset pairs are Cityscapes to Foggy Cityscapes, Sim10k to Cityscapes, KITTI to Cityscapes, and Pascal VOC to Clipart. Additionally, specialized scenarios such as BDD100K daytime to BDD100K nighttime are employed for specific applications. In the following, we provide an independent review of these datasets and conclude with a comparative analysis of model performance across widely utilized dataset benchmarks.

**Cityscape**(Cordts et al. 2016)**:** Each image in this dataset has precise labels for various objects present in the image. These labels include object categorization, object boundary detection, key point detection on objects, and other related information. It serves as one of the benchmark datasets in road/urban scene semantic segmentation and includes high-resolution images (1080p) of various urban environments collected from different parts of various cities.

**Foggy Cityscapes**(Sakaridis et al. 2018)**:** The main feature of this dataset is the presence of fog and dust in the images, which can pose a challenge for deep learning models. Fog and dust usually reduce the clarity and quality of images, making object and feature detection typically more difficult. This dataset includes images of urban scenes in foggy and dusty weather conditions.

**Caltech**(Fei-Fei et al. 2004)**:** This dataset is designed for the development and evaluation of object detection algorithms and models in the field of machine vision and deep learning.

**SYNTHIA**(Ros et al. 2016)**:** This dataset is created to provide a real-world test and simulation environment for analyzing machine learning algorithms and models under various urban and vehicular imaging conditions. It contains images from a virtual city in 13 classes with artificial lighting in both day and night conditions. It also has good environmental diversity and a relatively high similarity to real images.

**KITTI**[16](Liao et al. 2022)**:** The KITTI dataset is one of the well-known datasets in the field of machine vision and image processing for various tasks, including object detection, depth estimation, object tracking, and more. KITTI is a commonly used dataset in autonomous driving research, but it only provides data under normal weather conditions.

**PascalVOC**(Dong et al. 2021)**:** This dataset includes common objects. It is introduced for diverse tasks such as large-scale categorization, object detection, and segmentation.

**SIM10K**(Johnson-Roberson et al. 2017)**:** This dataset includes annotations for object detection and semantic segmentation tasks, making it suitable for training deep learning networks.

**BDD100K**(Yu et al. 2020)**:** This dataset is currently the largest dataset for autonomous driving AI. It is also designed to facilitate algorithmic studies on large and diverse image data and multiple tasks. This dataset includes various tasks such as object detection, instance segmentation, semantic segmentation, lane detection, drivable area segmentation, and more. It is suitable for training models that can perform a combination of perceptual tasks with different complexities instead of performing only identical tasks with the same prediction structure.

**YT-BB**[17](Real et al. 2017)**:** YTBB, a collection of video addresses with labeled bounding boxes for objects with a large data frequency, is developed for object detection issues in YouTube videos.

**Clipart**(Inoue et al. 2018)**:** The Clipart1k dataset is a collection of images used for various object detection tasks, including weakly supervised and unsupervised object detection. The images in this dataset are collected from sources like CMPlaces, Openclipart2, and Pixabay3, and they feature a variety of objects and scenes with complex backgrounds.

**Watercolor**(Inoue et al. 2018)**:** This dataset is primarily used for cross-domain object detection tasks, which involve training models to recognize objects across different visual domains. It is particularly useful for research in weakly supervised and unsupervised object detection. The Watercolor2k dataset is often used alongside other datasets like Clipart1k and Comic2k to evaluate the performance of object detection models in different artistic styles.

**DomainNet**(Peng et al. 2019): DomainNet dataset is one of the largest and most diverse datasets designed specifically for domain adaptation research. This dataset addresses the challenge of multi-source domain adaptation by providing a comprehensive benchmark that encompasses both inter- and intra-domain variations.

**Synscapes**(Wrenninge and Unger 2018): Synscapes is a synthetic dataset designed for autonomous driving, offering greater diversity for evaluating our approach. It includes a collection of 25,000 training images.

In **Fig. 11**, some examples from these datasets are presented.

[16] KITTI Vision Benchmark Suite

[17] YouTube Bounding Boxes

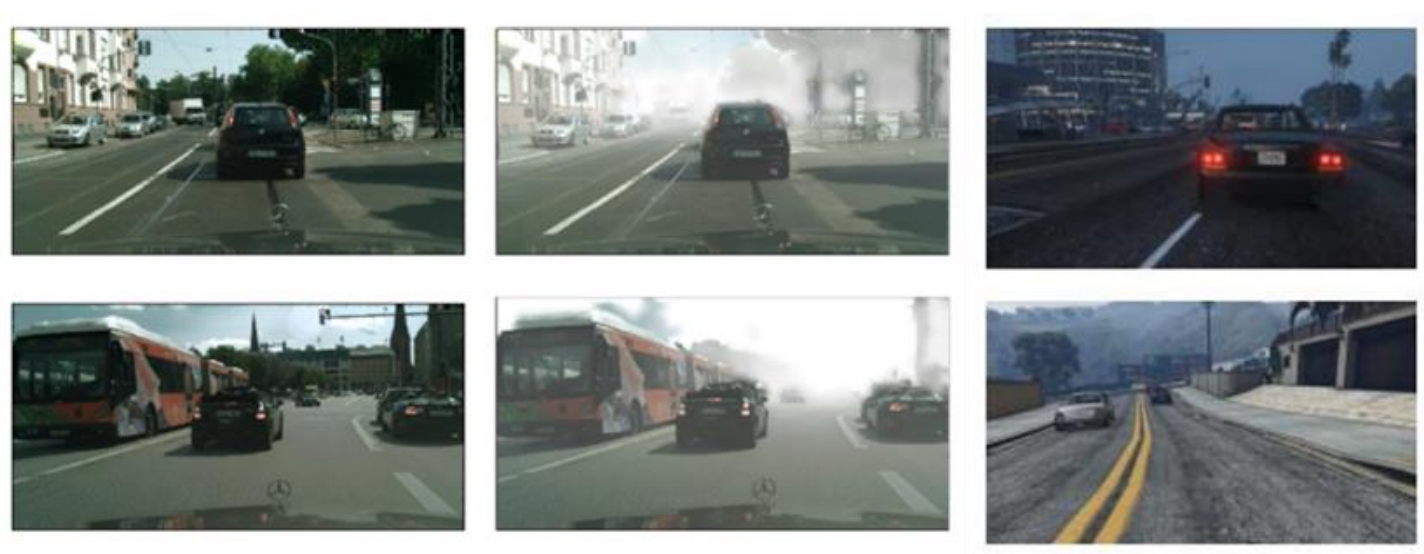

**Fig. 11.** Representative samples from the datasets used in this study (from left to right): (a) Cityscapes, (b) Foggy Cityscapes and (c) Sim10k.

**Table 2** provides a comprehensive overview of key datasets used in computer vision and domain adaptation for object detection tasks research. The table outlines essential attributes, including the primary use case, environmental conditions, whether the dataset is synthetic or real, the number of images, resolution, dataset name, and reference number. It also facilitates comparisons in terms of scale, resolution, and the contexts in which the datasets are utilized, providing a clear understanding of their relevance to specific research objectives.

**Table 2.** Table of compression between datasets that use in domain adaptation for object detection tasks

| Num | Dataset name | Resolution | Number of Images | Synthetic or Real | Environmental Conditions | Primary Use Case |
|---|---|---|---|---|---|---|
| 1 | **Cityscape** (Cordts et al. 2016) | 1080p | 5000 | Real | Urban Environments | Urban scene semantic segmentation |
| 2 | **Foggy Cityscapes** (Sakaridis et al. 2018) | 2048x1024 | 8000 | Real/ Synthetic | Foggy, Dusty | Object detection in adverse weather |
| 3 | **Caltech** (Fei-Fei et al. 2004) | Variable | 250000 | Real | Various: Urban scenes with pedestrians, vehicles, and varying weather and lighting | Pedestrian detection |
| 4 | **SYNTHIA** (Ros et al. 2016) | Variable | 200000 | Synthetic | Day & Night | Autonomous driving research |
| 5 | **KITTI** (Liao et al. 2022) | 1242x375 | 15000 | Real | Urban Environments | Autonomous vehicle perception |
| 6 | **PascalVOC** (Dong et al. 2021) | Variable | 11530 | Real | Various: Indoor and outdoor scenes, multiple object types, varied backgrounds | Object categorization and segmentation |
| 7 | **SIM10K**(Johnson-Roberson et al. 2017) | Variable | 10000 | Synthetic | Various: Synthetic urban environments with varying vehicle types and road scenes | Vehicle detection in synthetic scenes |
| 8 | **BDD100K** (Yu et al. 2020) | 1280x720 | 100000 | Real | Various: Day & night, urban and rural roads, diverse weather (clear, rain, fog, snow) | Large-scale autonomous driving research |
| 9 | **YT-BB**(Real et al. 2017) | Variable | 380000 | Real | Various: Video data with different objects, scenes, and lighting conditions | Object detection in video data |
| 10 | **Clipart**(Inoue et al. 2018) | Variable | 1000 | Synthetic | Various: Artistic and stylized images with diverse object representations and backgrounds | Domain adaptation research |
| 11 | **Watercolor**(Inoue et al. 2018) | Variable | 1000 | Synthetic | Various (Artistic): Artistic scenes with watercolor-style images, diverse object contexts | Synthetic-to-real domain adaptation |
| 12 | **DomainNet**(Peng et al. 2019) | Variable | 600000 | Real/Synthetic[18] | Multiple Domains: Real-world, clipart, sketch, and painting images, diverse environments | Synthetic-to-real domain adaptation |

## 5 Discussion about Research Progress

This section presents a concise summary of the key advancements in domain adaptation for object detection. It first offers a timeline that tracks the evolution of significant methods and milestones in the field. Following this, a comparative table is provided, which categorizes and contrasts the various approaches based on their techniques, datasets, and performance metrics. This overview sets the stage for the concluding discussion, highlighting the progress made and identifying future directions for research in domain adaptation for object detection.

**Fig. 12** illustrates the publication trends in the domain of "Domain Adaptation for Object Detection." The data is derived from Google Scholar and represents the annual growth in research contributions from 2018 to 2024. This upward trend highlights the increasing interest and significance of this topic in the research community.

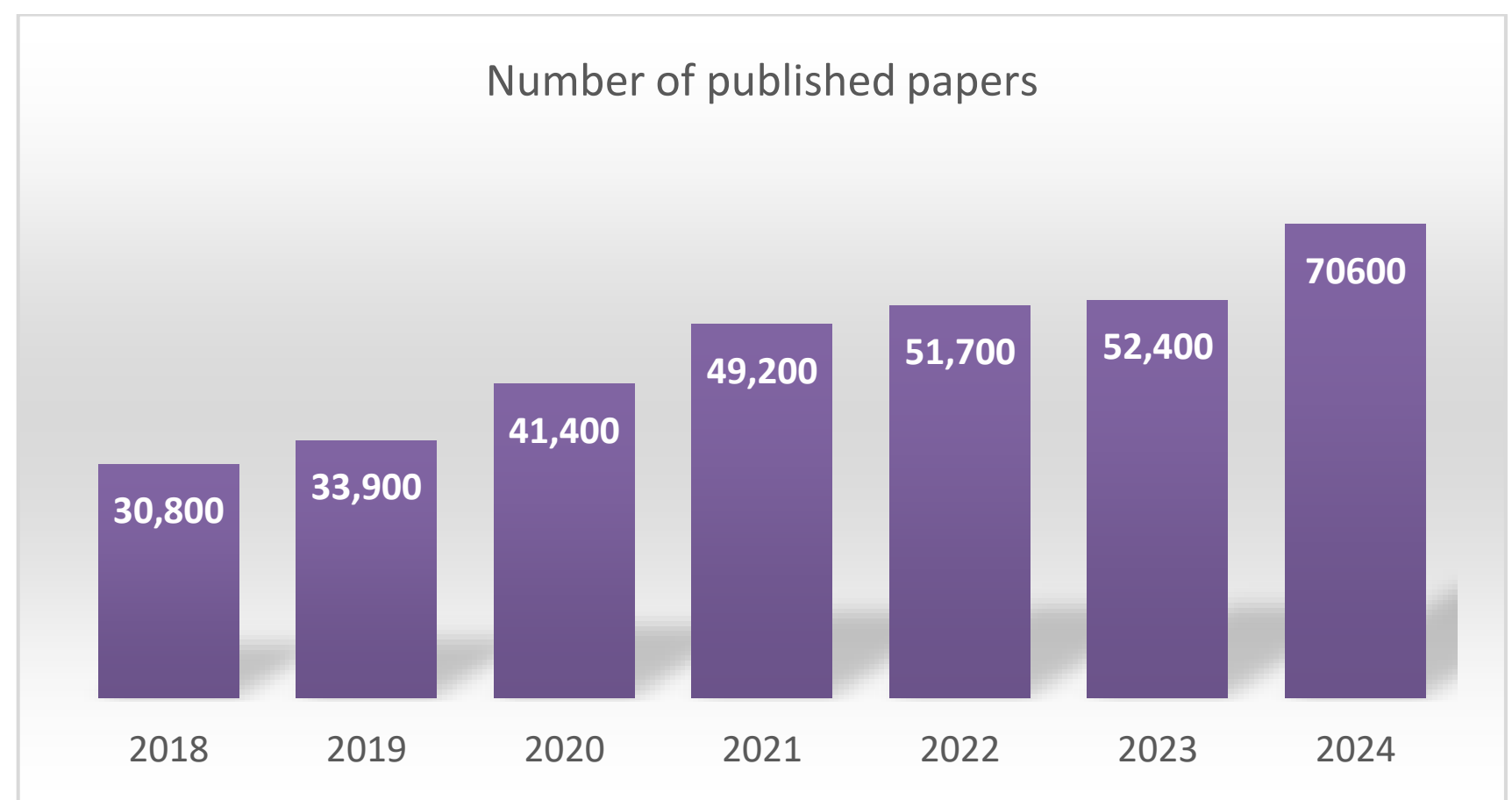

**Fig. 12.** Publication trends in "Domain Adaptation for Object Detection" (2018–2024) based on Google Scholar data. The data reveals a steady increase in publications, reflecting the growing interest and advancements in this research area.

The timeline illustrated in **Fig. 13** provides a comprehensive chronological overview of advancements in domain adaptation research for object detection spanning the years 2018 to 2024. This visual representation emphasizes key milestones and influential publications that have shaped the trajectory of the field.

Specifically, some of the most highly cited works from 2018 and 2019, denoted as DA-Faster (Chen et al. 2018), ZDDA (Peng et al. 2018), Cross-domain-detection (Inoue et al. 2018), Diversify and Match (Kim et al. 2019), Noisy Labeling (Khodabandeh et al. 2019), Automatic adaptation (RoyChowdhury et al. 2019) and Selective Cross-Domain Alignment (Zhu et al. 2019) have been included to highlight foundational contributions during the early stages of this research domain. These studies laid the groundwork for subsequent innovations, introducing critical concepts and methodologies. Similarly, the timeline features prominent publications from 2020 and 2021, identified as Progressive Domain Adaptation (Hsu et al. 2020), CST_DA_detection (Zhao et al. 2020), SW-Faster-ICR-CCR (Xu et al. 2020), coarse-to-fine feature adaptation (Zheng et al. 2020) , ST3D (Yang et al. 2021), SPG (Xu et al. 2021), SimROD (Ramamonjison et al. 2021), C2FDA (Zhang et al. 2022) and UaDAN (Guan et al. 2022), which reflect the significant progress achieved during this period, including the development of advanced frameworks and techniques that further enhanced performance in domain adaptation tasks.

[18] This dataset includes synthetic images like clipart and sketches, along with real photos.

Moreover, the timeline also incorporates several state-of-the-art approaches from 2022 to 2024, showcasing the cutting-edge methods and innovations that are currently defining the field. These recent contributions not only demonstrate remarkable performance improvements but also highlight emerging trends and directions for future research. By capturing these pivotal works, the timeline provides a holistic perspective on the evolution and current landscape of domain adaptation for object detection.

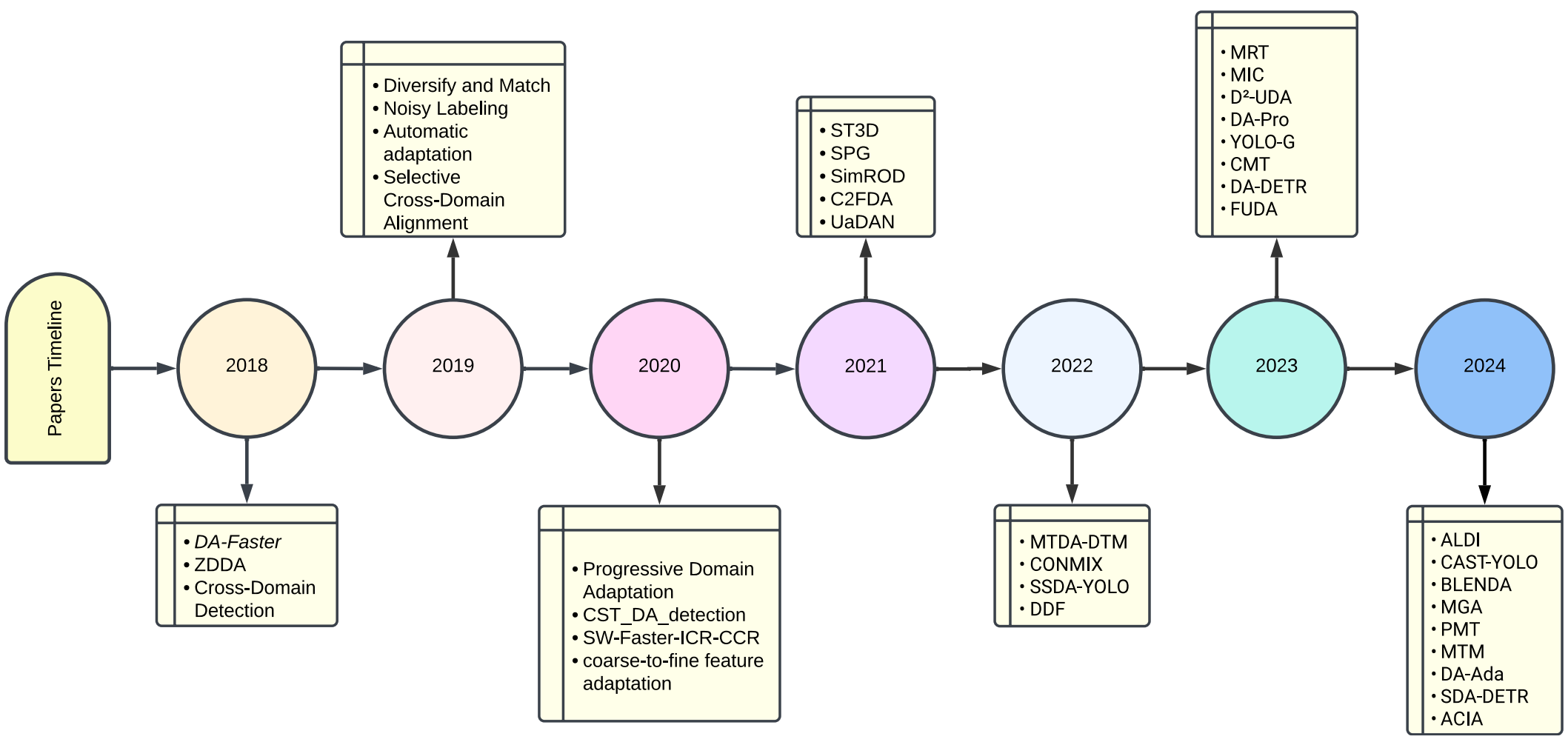

**Fig. 13.** An Overview of Key Advancements in Domain Adaptation for Object Detection (2018-2024)

**Table 3** offers a detailed comparison of state-of-the-art domain adaptation methods for object detection, summarizing their key characteristics, methodologies, and performance across various datasets. The table includes an overview of the object detectors utilized in each study. It also highlights the domain adaptation strategies employed, which are categorized into approaches like adversarial learning, discrepancy minimization, teacher-student frameworks, multi-domain methods, ensemble models and VLMs.

The corresponding papers for each method are further discussed in Section 2.3 and Section 2.4, where we provide detailed insights into each study, their contributions, and their impact on domain adaptation in object detection. This comparative analysis not only demonstrates the diversity of approaches but also provides a clear benchmark for assessing advancements in domain adaptation for object detection, serving as a valuable resource for researchers in this field.

**Table 3.** Table of comparison between articles

| Num | Papers | Year | Detector | Feature alignment method | | | | Domain Adaptation Method | | | | | | Synthetic data | dataset | mAP |
|---|---|---|---|---|---|---|---|---|---|---|---|---|---|---|---|---|
| | | | | Feature Reconstruction/Augmentation | Feature alignment | | Feature transformation | Adversarial | discrepancy | Teacher-Student | Ensemble | Multi-Domain | VLM | | | |
| | | | | | Image level | Instance level | | | | | | | | | | |
| 1 | ALDI (Kay et al. 2024) | 2024 | Faster R-CNN | | ✓ | ✓ | | ✓ | ✓ | ✓ | | | | ✓ | Cityscapes → Foggy Cityscapes<br>Sim10k → Cityscapes | 66.8<br>78.2 |
| 2 | CAST-YOLO (Liu et al. 2023b) | 2024 | YOLO | | ✓ | | ✓ | | | ✓ | | | | × | Cityscapes → Foggy Cityscapes | 43.3 |
| 3 | BlenDA (Huang et al. 2024) | 2024 | Transformer | | | ✓[19] | ✓ | ✓ | | | | | | × | Cityscapes → Foggy Cityscapes<br>Cityscapes → BDD100k-daytime | 53.4<br>33.5 |
| 4 | MGA (Zhang et al. 2024b) | 2024 | Faster R-CNN | | ✓ | ✓ | | ✓ | | ✓ | | | | ✓ | Cityscapes → Foggy Cityscapes<br>PascalVOC → Clipart<br>PascalVOC → Watercolor<br>Sim10k → Cityscapes | 47.4<br>47<br>62.1<br>54.3/ 55.5 |
| 5 | MTM (Weng and Yuan 2024) | 2024 | Transformer | ✓ | ✓ | ✓ | | ✓ | | ✓ | | | | ✓ | Cityscapes → Foggy Cityscapes<br>Cityscapes → BDD100k<br>Sim10k → Cityscapes | 48.9<br>37.3<br>58.1 |
| 6 | DA-Ada (Li et al. 2024) | 2024 | Transformer | | ✓ | ✓ | | ✓ | | | | | ✓ | ✓ | Cityscapes → Foggy Cityscapes<br>KITTI → Cityscape<br>Sim10k → Cityscapes<br>PascalVOC → Clipart | 58.5<br>66.7<br>67.3<br>48.0 |
| 7 | PMT (Belal et al. 2024a) | 2024 | Faster R-CNN | | ✓ | ✓ | | ✓ | | ✓ | | ✓ | | ✓ | BDD100k-daytime (S1) + night time (S2) → BDD100k-dusk/down<br><br>Cityscape (S1) + KITTI (S2) → BDD100k-daytime<br><br>Cityscapes (S1) + MS COCO (Lin et al. 2014) (S2) + Synscapes (S3) → BDD100k-daytime | 45.3<br><br>58.7<br><br>39.7 |

[19] Also have channel align

| | | | | | | | | | | | | | | | | |
|---|---|---|---|---|---|---|---|---|---|---|---|---|---|---|---|---|
| 8 | SDA-DETR (Zhang et al. 2024a) | 2024 | Transformer | | ✓ | ✓ | | ✓ | ✓ | | | | | × | Cityscapes → Foggy Cityscapes | 47.3 |
| 9 | ACIA(Belal et al. 2024b) | 2024 | Faster R-CNN | | ✓ | ✓ | | ✓ | | ✓ | | ✓ | | ✓ | BDD100k-daytime (S1) + night time (S2) → BDD100k-dusk/down<br><br>Cityscape (S1) + KITTI (S2) → BDD100k-daytime<br><br>Cityscapes (S1) + MS COCO (Lin et al. 2014) (S2) + Synscapes (S3) → BDD100k-daytime | 47.9<br><br>59.1<br><br>42.3 |
| 10 | $D^2$-UDA (Zhu et al. 2023a) | 2023 | Faster R-CNN | | | ✓ | | ✓ | | ✓ | | | | ✓ | Cityscapes → Foggy Cityscapes<br>Sim10k → Cityscapes | 53.5<br>58.1 |
| 11 | DA-Pro (Li et al. 2023) | 2023 | Transformer | | ✓ | | | ✓ | | | ✓ | | ✓ | ✓ | Cityscapes → Foggy Cityscapes<br>KITTI → Cityscape<br>Sim10k → Cityscapes | 55.9<br>61.4<br>62.9 |
| 12 | MRT (Zhao et al. 2023) | 2023 | Transformer | ✓ | | ✓ | ✓ | ✓ | | ✓ | | | | ✓ | Cityscapes → Foggy Cityscapes<br>Cityscapes → BDD100k-daytime<br>Sim10k → City-scapes(car) | 51.2<br>33.7<br>62.0 |
| 13 | MIC (Hoyer et al. 2023) | 2023 | Faster R-CNN | ✓ | | | | ✓ | | ✓ | | | | × | Cityscapes → Foggy Cityscapes | 47.6 |
| 14 | YOLO-G (Wei et al. 2023) | 2023 | YOLO | | | ✓ | | ✓ | | | | | | ✓ | Cityscapes → Foggy Cityscapes<br>Sim10k → Cityscapes<br>KITTI → Cityscape<br>PascalVOC → Clipart<br>Pascal VOC → Water-color<br>Cityscapes → BDD100k | 47.8<br>64.2<br>62.8<br>44.3<br>53.1<br>34.6 |
| 15 | CMT (Cao et al. 2023) | 2023 | Faster R-CNN | | | ✓ | | | | ✓ | | | | ✓ | Cityscapes → Foggy Cityscapes<br>PascalVOC → Clipart<br>KITTI → Cityscape | 51.9<br>47.0<br>64.3 |
| 16 | DA-DETR (Zhang et al. 2023) | 2023 | Transformer | | ✓ | ✓ | | ✓ | | | ✓ | | | ✓ | Cityscapes → Foggy Cityscapes<br>Sim10k → Cityscapes<br>KITTI → Cityscape<br>PascalVOC → Clipart<br>Pascal VOC → Water-color2k<br>Pascal VOC → Comic2k (Hsu et al. 2020) | 43.5<br>54.7<br>48.9<br>41.3<br>50.6<br>35.1 |
| 17 | FUDA (Zhu et al. 2023b) | 2023 | Faster R-CNN | | ✓ | ✓ | | ✓ | ✓ | ✓ | | | | ✓ | Cityscapes → Foggy Cityscapes<br>Cityscapes → Rainy Cityscapes<br>KITTI → KITTI Rainy | 40.2<br>47.9<br>53.9 |
| 18 | SSDA-YOLO (Zhou et al. 2023) | 2022 | YOLO | | ✓ | ✓ | | ✓ | ✓ | ✓ | | | | ✓ | PascalVOC → Clipart<br>Cityscapes → Foggy Cityscapes | 44.3<br>55.9 |
| 19 | MTDA-DTM (Nguyen-Meidine et al. 2022) | 2022 | Transformer | | ✓ | ✓ | | ✓ | ✓ | ✓ | | ✓ | | ✓ | Cityscapes → Foggy Cityscapes (T1) - → Rainy Cityscapes (T2)<br>PascalVOC → Clipart (T1) → Watercolor (T2)<br>PascalVOC → Clipart (T1) → Watercolor (T2) → Comic(T3) | 39.1<br>38.9<br>38.1 |
| 20 | DDF(Liu et al. 2023a) | 2022 | Faster R-CNN | ✓ | | | | ✓ | ✓ | | | | | ✓ | Cityscapes → Foggy Cityscapes<br>KITTI → Cityscape<br>Cityscape → KITTI<br>Sim10k → Cityscapes | 42.3<br>46.0<br>75.0<br>44.3 |

## 6 Conclusion

In this review, we have explored the evolution and current state of feature-based domain adaptation methods for object detection. As highlighted, domain adaptation plays a crucial role in enabling models to generalize effectively when faced with domain shifts, a challenge that is common in real-world applications such as autonomous driving, medical imaging, and surveillance. By categorizing recent advancements into six primary strategies—discrepancy methods, adversarial methods, multi-domain methods, teacher-student methods, ensemble methods, and vision-language models—we have provided a comprehensive overview of how each approach addresses the key issues of domain misalignment and data scarcity. While substantial progress has been made, several challenges remain, such as mitigating negative transfer, handling noisy labels, and ensuring scalable solutions for large, diverse datasets. Advancements like adversarial training and teacher-student frameworks have demonstrated their potential in improving model robustness and performance, yet the methods still require refinement to handle larger domain shifts and more complex environments. The integration of multi-domain adaptation techniques and vision-language models also shows promising results, offering new avenues for cross-domain generalization and robust object detection in dynamic settings. Moving forward, future research should focus on developing hybrid models that combine the strengths of these various methods. The continued evolution of deep learning architectures, along with more efficient training techniques, could unlock even greater potential for domain adaptation in object detection. Moreover, as real-world applications demand real-time adaptation, methods that enable online domain adaptation will become increasingly important. This combination of theoretical and practical advancements is essential for achieving truly adaptable and reliable object detection systems.

In conclusion, feature-based domain adaptation remains a dynamic and critical area of research in machine learning, with broad implications for various industries. By addressing the current limitations and building upon recent advancements, future work has the potential to make significant contributions toward developing universally adaptable object detection models.